\documentclass[10pt,twocolumn,letterpaper]{article}

\usepackage{cvpr}
\usepackage{times}
\usepackage{epsfig}
\usepackage{graphicx}
\usepackage{amsmath}
\usepackage{amssymb}

\usepackage{multirow, subcaption}
\usepackage[linesnumbered,ruled,vlined]{algorithm2e}

\SetKwInput{KwInput}{Input}  
\SetKwInput{KwOutput}{Output} 

\usepackage[pagebackref=true,breaklinks=true,letterpaper=true,colorlinks,bookmarks=false]{hyperref}

\cvprfinalcopy 

\def\cvprPaperID{5375}

\ifcvprfinal\fi
\begin{document}

\title{Meta-Transfer Learning for Zero-Shot Super-Resolution}

\author{Jae Woong Soh \qquad Sunwoo Cho \qquad Nam Ik Cho\\
Department of ECE, INMC, Seoul National University, Seoul, Korea\\
{\tt\small \{soh90815, etoo33\}@ispl.snu.ac.kr, nicho@snu.ac.kr}}

\maketitle
\thispagestyle{empty}

\begin{abstract}
Convolutional neural networks (CNNs) have shown dramatic improvements in single image super-resolution (SISR) by using large-scale external samples. Despite their remarkable performance based on the external dataset, they cannot exploit internal information within a specific image. Another problem is that they are applicable only to the specific condition of data that they are supervised. For instance, the low-resolution (LR) image should be a  ``bicubic'' downsampled noise-free image from a high-resolution (HR) one. To address both issues, zero-shot super-resolution (ZSSR) has been proposed for flexible internal learning. However, they require thousands of gradient updates, \ie, long inference time. In this paper, we present Meta-Transfer Learning for Zero-Shot Super-Resolution (MZSR), which leverages ZSSR. Precisely, it is based on finding a generic initial parameter that is suitable for internal learning. Thus, we can exploit both external and internal information, where \textbf{one single gradient update} can yield quite considerable results. (See \figurename{~\ref{fig:001}}). With our method, the network can quickly adapt to a given image condition. In this respect, our method can be applied to a large spectrum of image conditions within a fast adaptation process.
\end{abstract}

\section{Introduction}

\begin{figure}[t]
	\begin{center}
		\captionsetup{justification=centering}
		
		\begin{subfigure}[t]{0.48\linewidth}
			\centering
			\includegraphics[width=1\columnwidth]{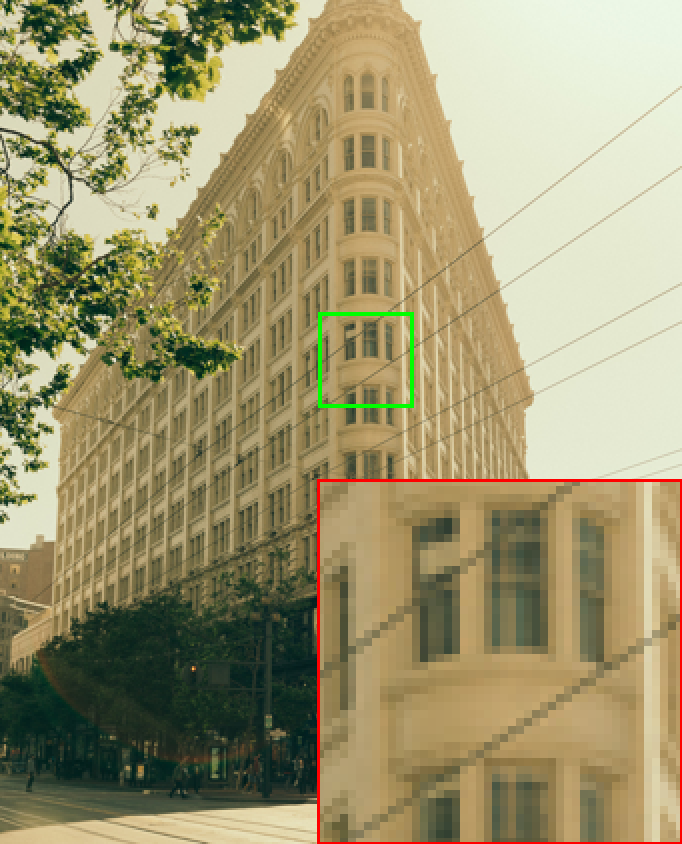}
			\caption{LR}
		\end{subfigure}
		\begin{subfigure}[t]{0.48\linewidth}
			\centering
			\includegraphics[width=1\columnwidth]{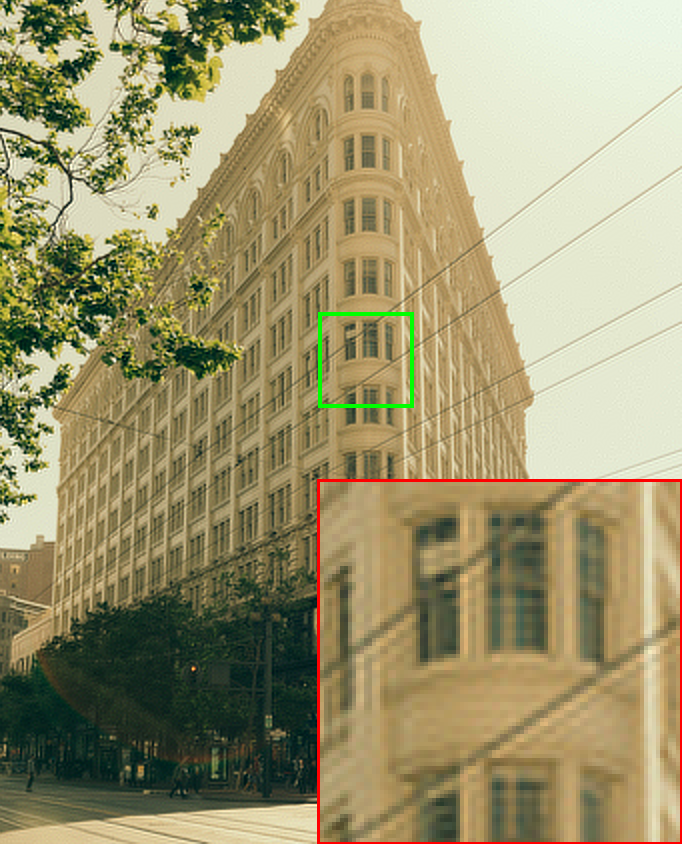}
			\caption{ZSSR \cite{ZSSR}\\ \emph{2,850 updates}}
		\end{subfigure}
		\begin{subfigure}[t]{0.48\linewidth}
			\centering
			\includegraphics[width=1\columnwidth]{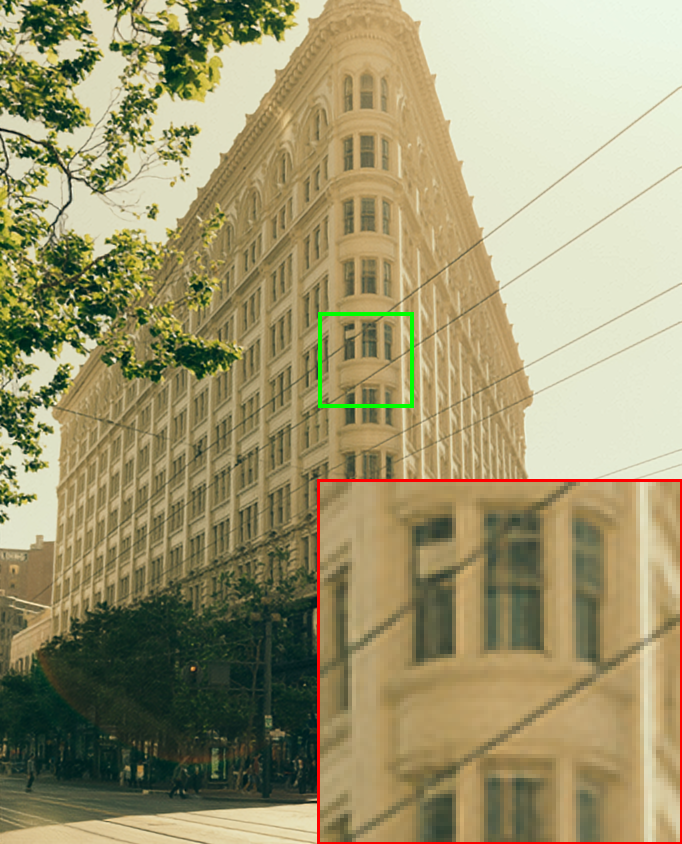}
			\caption{Fine-tuning \\ \emph{2,000 updates} }
		\end{subfigure}
		\begin{subfigure}[t]{0.48\linewidth}
			\centering
			\includegraphics[width=1\columnwidth]{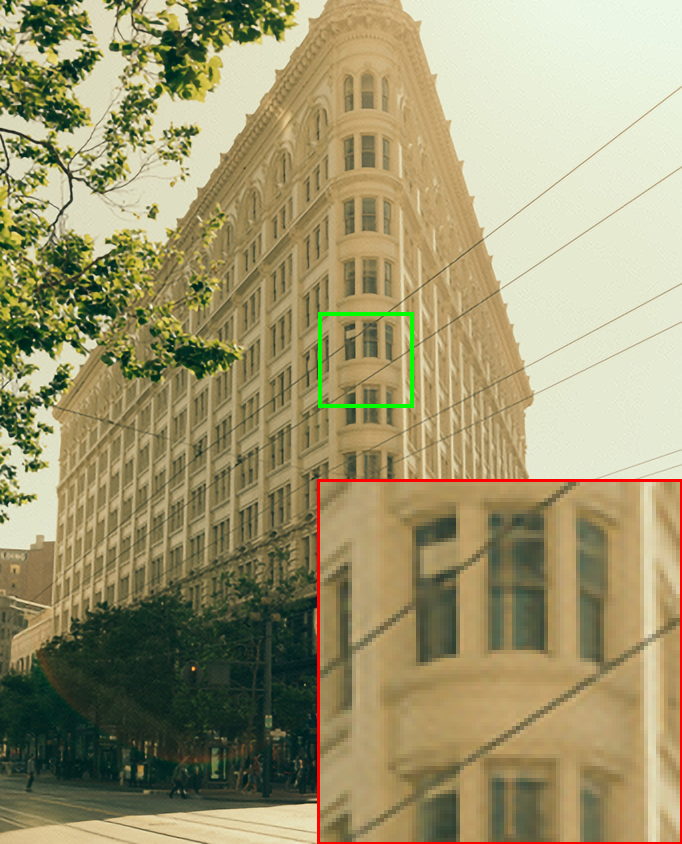}
			\caption{ MZSR (Ours) \\ \emph{\textbf{One} update} }
		\end{subfigure}		

	\end{center}
	\caption{Super-resolved results ($\times 2$) of ``img050'' in Urban100 \cite{Self-Exemplar}. The blur kernel of the LR image is an isotropic Gaussian kernel with width $2.0$. Result of (c) is fine-tuned from a pre-trained model. Our MZSR outperforms other methods within just \emph{one} single gradient descent update.}
	\label{fig:001}
\end{figure}

SISR, which is to find a plausible HR image from its counterpart LR image, is a long-standing problem in low-level vision area. Recently, the remarkable success of CNNs brought attention to the research community, and hence numerous CNN-based SISR methods have exhibited large performance leap~\cite{VDSR, SRGAN, EDSR, RDN, CARN, RCAN, NatSR, SRFBN, DBPN, OISR}. Most of the recent state-of-the-art (SotA) CNN-based methods are based on a large number of external training dataset and self-supervised settings with \emph{known} degradation model, \eg, ``bicubic'' downsampling. Impressively, the recent SotA CNNs show significant PSNR gains compared to the conventional large size of models for the noise-free ``bicubic'' downsampling condition. However, in real-world situations, when the LR image has distant statistics in downsampling kernels and noises, the recent methods produce undesirable artifacts and show inferior results due to the domain gap. Moreover, their number of parameters and memory overheads are usually too large to be used in real applications.


Besides, non-local self-similarity in scale and across multi-scale, which is the internal recurrence of information within a single image, is one of the strong natural image priors. Therefore it has long been used in image restoration tasks, including image denoising \cite{NLM, BM3D} and super-resolution \cite{Non-Param, Self-Exemplar}. Additionally, the powerful image prior of non-local property is embedded into network architecture \cite{Non-local, NLRN, RNAN} by implicitly learning such priors to boost the performance of the networks further. Also, some works to learn internal distribution have been proposed \cite{ZSSR, SinGAN, InGAN}. Moreover, there have been many studies to combine the advantages of external and internal information for image restoration \cite{comb1, comb2, comb3, comb4}.

Recently, ZSSR \cite{ZSSR} has been proposed for zero-shot super-resolution, which is based on the zero-shot setting to exploit the power of CNN but can be easily adapted to the test image condition. Interestingly, ZSSR learns the internal non-local structure of the test image, \ie, deep internal learning. Thus it outperforms external-based CNNs in some regions where the recurrences are salient. Also, ZSSR is highly flexible that it can address any blur kernels, and thus easily adapted to the conditions of test images.

However, ZSSR has a few limitations. First, it requires thousands of backpropagation gradient updates at test time, which requires considerable time to get the result. Also, it cannot fully exploit the large-scale external dataset, and rather it depends only on internal structure and patterns, which lacks in the number of total examples. Eventually, this leads to inferior results in most of the regions with general patterns compared to the external-based methods.

On the other hand, meta-learning or learning to learn fast has recently attracted many researchers. Meta-learning aims to address a problem that artificial intelligence is hard to learn new concepts quickly with a few examples, unlike human intelligence.
In this respect, meta-learning is jointly merged with few-shot learning, and many methods with this approach have been proposed \cite{Prototypical, Matching, Relation, TADAM, SNAIL, MAML, GRAD1, GRAD2, MTL}. Among them, Model-Agnostic Meta-Learning (MAML) \cite{MAML} has shown great impact, showing SotA performance by learning the optimal initial state of the model such that the base-learner can fast adapt to a new task within a few gradient steps. MAML employs the gradient update as meta-learner, and the same author analyzed that gradient descent can approximate any learning algorithm \cite{Meta_Grad}. Moreover, Sun \etal \cite{MTL} have jointly utilized MAML with transfer learning to exploit large-scale data for few-shot learning.

Inspired by the above-stated works and ZSSR, we present Meta-Transfer Learning for Zero-Shot Super-Resolution (MZSR), which is kernel-agnostic. We found that simply employing transfer learning or fine-tuning from a pre-trained network does not yield plausible results. As ZSSR only has a meta-test step, we additionally adopt a meta-training step to make the model adapt fast to new blur kernel scenarios. Additionally, we adopt transfer learning in advance to fully utilize external samples, further leveraging the performance.
In particular, transfer learning with the help of a large-scale synthetic dataset (``bicubic'' degradation setting) is first performed for the external learning of natural image priors. Then, meta-learning plays a role in learning task-level knowledge with different downsampling kernels as different tasks. At the meta-test step, simple self-supervised learning is conducted to learn image-specific information within a few gradient steps. As a result, we can exploit both external and internal information. Also, by leveraging the advantages of ZSSR, we may use a lightweight network, which is flexible to different degradation conditions of LR images. Furthermore, our method is much faster than ZSSR, \ie, it quickly adapts to new tasks within a few gradient steps, while ZSSR requires thousands of updates.

In summary, our overall contribution is three-fold:
\begin{itemize}
	\item We present a novel training scheme based on meta-transfer learning, which learns an effective initial weight for fast adaptation to new tasks with the zero-shot unsupervised setting.
	\item By using external and internal samples, it is possible to leverage the advantages of both internal and external learning.
	\item Our method is fast, flexible, lightweight and unsupervised at meta-test time, hence, eventually can be applied to real-world scenarios.
\end{itemize}

\begin{figure*}[t]
	\begin{center}
		\centering
		\includegraphics[width=0.8\linewidth]{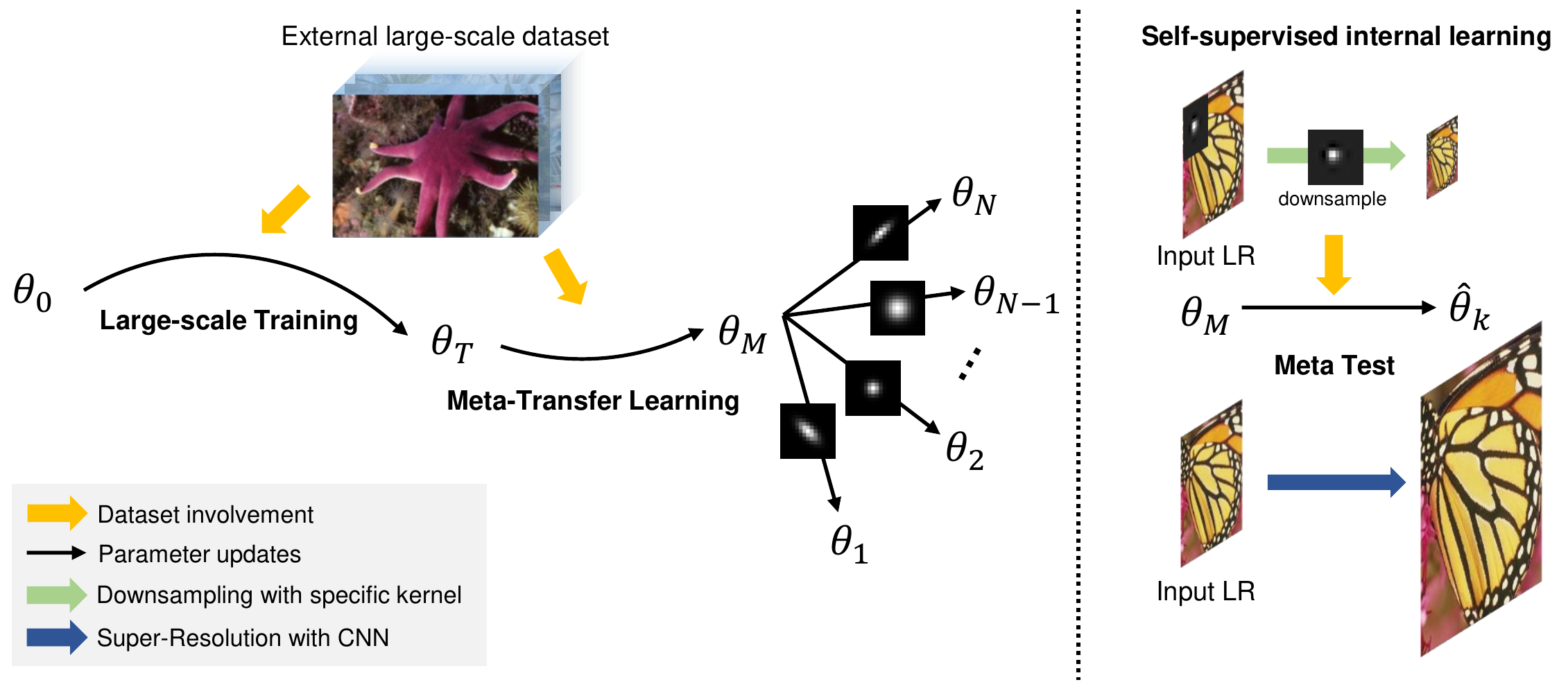}
	\end{center}
	\caption{The overall scheme of our proposed MZSR. During meta-transfer learning, the external dataset is used, where internal learning is done during meta-test time. From random initial point $\theta_0$, large-scale dataset DIV2K \cite{DIV2K} with ``bicubic'' degradation is exploited to obtain $\theta_{T}$. Then, meta-transfer learning learns a good representation $\theta_{M}$ for super-resolution tasks with diverse blur kernel scenarios. The figure shows $N$ tasks for simplicity. In the meta-test phase, self-supervision within a test image is exploited to train the model with corresponding blur kernel.}
	\label{fig:overall}
\end{figure*}

\section{Related Work}

\subsection{CNN-based Super-Resolution}
SISR is based on the image degradation model as
\begin{equation}
\mathbf{I}^\mathbf{k}_{LR}=(\mathbf{I}_{HR} \ast \mathbf{k}) \downarrow_{s} + \mathbf{n},
\label{eq:1}
\end{equation}
where $\mathbf{I}_{HR}$, $\mathbf{I}^\mathbf{k}_{LR}$, $\mathbf{k}$,  $\ast$, $\downarrow_{s}$, and $\mathbf{n}$ denote HR, LR image, blur kernel, convolution, decimation with scaling factor of $s$, and white Gaussian noise, respectively. It is notable that diverse degraded conditions can be found in real-world scenes, with various unknown $\mathbf{k}$, $\downarrow_{s}$, and $\mathbf{n}$.

Recently, numerous CNN-based networks have been proposed to super-resolve LR image with \emph{known} downsampling kernel \cite{VDSR, SRGAN, EDSR, DBPN, RDN, CARN, NatSR, SRFBN, OISR}. They show extreme performances in ``bicubic'' downsampling scenarios but suffer in non-bicubic cases due to the domain gap. To cope with multiple degradation kernels, SRMD \cite{SRMD} has been proposed. With additional inputs of kernel and noise information, SRMD outperforms other SISR methods in non-bicubic conditions. Also, IKC~\cite{BlindSR} has been proposed for blind super-resolution. On the other hand, ZSSR~\cite{ZSSR} has been proposed to learn image specific internal structure with CNN, and has shown that it can be applied to real-world scenes due to its flexibility.

\subsection{Meta-Learning}
In recent years, diverse meta-learning algorithms have been proposed. They can be categorized into three groups. The first group is \emph{metric based methods} \cite{Prototypical, Relation, Matching}, which is to learn metric space in which learning is efficient within a few samples. The second group is \emph{memory network-based methods} \cite{memory, TADAM, SNAIL}, where the network learns across task knowledges and well generalizes to unseen tasks. The last group is \emph{optimization based methods}, where gradient descent plays a role as a meta-learner optimization \cite{GRAD1, GRAD2, Meta_Grad, MAML}. Among them, MAML \cite{MAML} has shown a great impact on the research community, and several variants have been proposed \cite{Reptile, MTL, MAML++, MAML_Embed}. MAML inherently requires second-order derivative terms, and the first-order algorithm has also been proposed in \cite{Reptile}. Also, to cope with the instability of MAML training, MAML++ \cite{MAML++} has been proposed. Moreover, MAML within embedded space has been proposed \cite{MAML_Embed}. In this paper, we employ MAML scheme for fast adaptation of zero-shot super-resolution.

\section{Preliminary}
We introduce self-supervised zero-shot super-resolution and meta-learning schemes with notations, following related works \cite{ZSSR, MAML}.
\paragraph{Zero-Shot Super-Resolution}
ZSSR \cite{ZSSR} is totally unsupervised or self-supervised. Two phases of training and test are both held in runtime. In training phase, the test image $\mathbf{I}_{LR}$ is downsampled with desired kernel to generate ``LR son'' denoted as $\mathbf{I}_{son}$, and $\mathbf{I}_{LR}$ becomes the HR supervision, ``HR father.'' Then, the CNN is trained with the LR-HR pairs generated by a single image. The training solely depends on the test image, thus learns specific internal information to given image statistics. In the test phase, the trained CNN then works as a feedforward network, and the test input image is fed to the CNN to get the super-resolved image $\mathbf{I}_{SR}$.

\paragraph{Meta-Learning}
Meta-learning has two phases: meta-training and meta-test. We consider a model $f_\theta (\cdot)$, which is parameterized by $\theta$, that maps inputs $\mathbf{x}$ to outputs $\mathbf{y}$. The goal of meta-training is to make the model to be able to adapt to a large number of different tasks. A task $\mathcal{T}_i$ is sampled from a task distribution $p(\mathcal{T})$ for meta-training. Within a task, training samples are used to optimize the base-learner with a task-specific loss $\mathcal{L}_{\mathcal{T}_{i}}$ and test samples are used to optimize the meta-learner. In meta-test phase, the model $f_\theta(\cdot)$ quickly adapts to a new task $\mathcal{T}_{new}$ with the help of meta-learner. MAML \cite{MAML} employs a simple gradient descent algorithm as the meta-learner and seeks to find an initial transferable point where a few gradient updates lead to a fast adaptation of the model to a new task.

In our case, the input $\mathbf{x}$ and the output $\mathbf{y}$ are $\mathbf{I}^\mathbf{k}_{LR}$ and $\mathbf{I}_{SR}$. Also, diverse blur kernels constitute the task distribution, where each task corresponds to the super-resolution of an image degraded by a specific blur kernel.

\section{Method}

The overall scheme of our proposed MZSR is shown in \figurename{~\ref{fig:overall}}. As shown, our method consists of three steps: large-scale training, meta-transfer learning, and meta-test.

\subsection{Large-scale Training}
\label{sec1}

This step is similar to the large-scale ImageNet \cite{ImageNet} pre-training for object recognition. In our case, we adopt DIV2K \cite{DIV2K}  which is a high-quality dataset $\mathcal{D}_{HR}$. Using \emph{known} ``bicubic'' degradation, we first synthesized large number of paired dataset $(\mathbf{I}_{HR}, \mathbf{I}^{bic}_{LR})$, denoted as $\mathcal{D}$. Then, we trained the network to learn super-resolution of ``bicubic'' degradation model by minimizing the loss,
\begin{equation}
\mathcal{L}^{\mathcal{D}}(\theta) = \mathbb{E}_{\mathcal{D} \sim (\mathbf{I}_{HR}, \mathbf{I}^{bic}_{LR})}[||\mathbf{I}_{HR} -f_\theta (\mathbf{I}^{bic}_{LR})||_1],
\label{eq:2}
\end{equation}
which is the pixel-wise L1 loss \cite{EDSR, ZSSR} between prediction and the ground-truth.

The large-scale training has contributions within two respects.
First, as super-resolution tasks share similar properties, it is possible to learn efficient representations that implicitly represent natural image priors of high-resolution images, thus making the network ease to be learned. Second, as MAML \cite{MAML} is known to show some unstable training, we ease the training phase of meta-learning with the help of well pre-trained feature representations.

\subsection{Meta-Transfer Learning}
\label{sec2}

Since ZSSR is trained with the gradient descent algorithm, it is possible to introduce an \emph{optimization-based} meta-training step with the help of gradient descent algorithm, which is proven to be a universal learning algorithm \cite{Meta_Grad}.

In this step, we seek to find a sensitive and transferable initial point of the parameter space where a few gradient updates lead to large performance improvements. Inspired by MAML, our algorithm mostly follows MAML but with several modifications.

Unlike MAML, we adopt different settings for meta-training and meta-test. In particular, we use the external dataset for meta-training, whereas internal learning is adopted for meta-test. This is because we intend our meta-learner to more focus on the kernel-agnostic property with the help of a large-scale external dataset.

We synthesize dataset for meta-transfer learning, denoted as $\mathcal{D}_{meta}$. $\mathcal{D}_{meta}$ consists of pairs, $(\mathbf{I}_{HR}, I^{\mathbf{k}}_{LR})$, with diverse kernel settings. Specifically, we used isotropic and anisotropic Gaussian kernels for the blur kernels. We consider a kernel distribution $p(\mathbf{k})$, where each kernel is determined by a covariance matrix $\mathbf{\Sigma}$. it is chosen to have a random angle $\Theta \sim U[0, \pi]$, and two random eigenvalues $\lambda_1 \sim U[1, 2.5 s]$, $\lambda_2 \sim U[1, \lambda_1]$ where $s$ denotes the scaling factor. Precisely, the covariance matrix is expressed as
\begin{align}
\mathbf{\Sigma}=
\begin{bmatrix} 
\cos(\Theta) & -\sin(\Theta)\\
\sin(\Theta) & \cos(\Theta)
\end{bmatrix}
\begin{bmatrix} 
\lambda_1 & 0\\
0 & \lambda_2
\end{bmatrix}
\begin{bmatrix} 
\cos(\Theta) & \sin(\Theta)\\
-\sin(\Theta) & \cos(\Theta)
\end{bmatrix}.
\label{eq:3}
\end{align}

Eventually, we train our meta-learner based on $\mathcal{D}_{meta}$. We may divide $\mathcal{D}_{meta}$ into two groups: $\mathcal{D}_{tr}$ for task-level training, and $\mathcal{D}_{te}$ for task-level test.

In our method, adaptation to a new task $\mathcal{T}_i$ with respect to the parameters $\theta$ is one or more gradient descent updates. For one gradient update, new adapted parameters $\theta_{i}$ is then
\begin{equation}
\theta_{i} = \theta - \alpha \nabla_{\theta} \mathcal{L}^{tr}_{\mathcal{T}_{i}}(\theta),
\end{equation}
where $\alpha$ is the task-level learning rate. The model parameters $\theta$ are optimized to achieve minimal test error of $\mathcal{D}_{meta}$ with respect to $\theta_{i}$. Concretely, the meta-objective is
\begin{align}
&\arg \min_{\theta} \sum_{\mathcal{T}_i\sim p(\mathcal{T})} \mathcal{L}^{te}_{\mathcal{T}_{i}}(\theta_{i})\\
=&\arg \min_{\theta} \sum_{\mathcal{T}_i\sim p(\mathcal{T})} \mathcal{L}^{te}_{\mathcal{T}_{i}}(\theta - \alpha \nabla_{\theta} \mathcal{L}^{tr}_{\mathcal{T}_{i}}(\theta)).
\label{eq:6}
\end{align}

Meta-transfer optimization is performed using Eq. \ref{eq:6}, which is to learn the knowledge across task. Any gradient-based optimization can be used for meta-transfer training. For stochastic gradient descents, the parameter update rule is expressed as
\begin{equation}
\theta \leftarrow \theta - \beta \nabla_{\theta} \sum_{\mathcal{T}_{i}\sim p(\mathcal{T})} \mathcal{L}^{te}_{\mathcal{T}_i} (\theta_{i}),
\end{equation}
where $\beta$ is the meta-learning rate.

\subsection{Meta-Test}

The meta-test step is exactly the zero-shot super-resolution. As evidence in \cite{ZSSR}, this step enables our model to learn internal information within a single image. With a given LR image, we downsample it with corresponding downsampling kernel (kernel estimation algorithms \cite{Non-Param, KernelEst} can be adopted for blind scenario) to generate $\mathbf{I}_{son}$ and perform a few gradient updates with respect to the model parameter using a single pair of ``LR son'' and a given image. Then, we feed a given LR image to the model to get a super-resolved image.

\begin{algorithm}
	\DontPrintSemicolon
	\SetAlgoLined
	
	\KwInput{High-resolution dataset $\mathcal{{D}_{HR}}$ and blur kernel distribution $p(\mathbf{k})$}
	\KwInput{$\alpha$, $\beta$: learning rates}
	\KwOutput{Model parameter $\theta_M$}
	Randomly initialize $\theta$\\
	Synthesize paired dataset $\mathcal{D}$ by bicubicly downsample $\mathcal{D}_{HR}$\\
	\While{not done}
	{
		Sample LR-HR batch from $\mathcal{D}$\\
		Evaluate $\mathcal{L}^{D}$ by Eq. \ref{eq:2}\\
		Update $\theta$ with respect to $\mathcal{L}^{D}$
		
	}
	Generate task distribution $p(\mathcal{T})$ with $\mathcal{D}_{HR}$ and $p(\mathbf{k})$\\
	\While{not done}
	{
		Sample task batch $\mathcal{T}_i \sim p(\mathcal{T})$\\
		\For{all $\mathcal{T}_{i}$}{
			Evaluate training loss ($\mathcal{D}_{tr}$): $\mathcal{L}^{tr}_{\mathcal{T}_{i}}(\theta)$ \\
			Compute adapted parameters with gradient descent: $\theta_i=\theta - \alpha \nabla_{\theta} \mathcal{L}^{tr}_{\mathcal{T}_{i}}(\theta)$
		}
		Update $\theta$ with respect to average test loss ($\mathcal{D}_{te}$):\\
		$\theta \leftarrow \theta - \beta \nabla_{\theta} \sum_{\mathcal{T}_{i}\sim p(\mathcal{T})} \mathcal{L}^{te}_{\mathcal{T}_{i}}(\theta_i)$
		
	}

	\caption{Meta-Transfer Learning}
	\label{alg:1}
\end{algorithm}

\begin{algorithm}
	\DontPrintSemicolon
	\SetAlgoLined
	
	\KwInput{LR test image $\mathbf{I}_{LR}$, meta-transfer trained model parameter $\theta_M$, number of gradient updates $n$ and learning rate $\alpha$}
	\KwOutput{Super-resolved image $\mathbf{I}_{SR}$}
	Initialize model parameter $\theta$ with $\theta_M$\\
	Generate LR son $\mathbf{I}_{son}$ by downsampling $\mathbf{I}_{LR}$ with corresponding blur kernel.\\
	\For{n steps}
	{
		Evaluate loss $\mathcal{L}(\theta)=||\mathbf{I}_{LR}-f_\theta (\mathbf{I}_{son})||_1$\\
		Update $\theta \leftarrow \theta - \alpha \nabla_{\theta}\mathcal{L}(\theta)$
		
	}
	\Return $\mathbf{I}_{SR}=f_\theta(\mathbf{I}_{LR})$
	
	\caption{Meta-Test}
	\label{alg:2}
\end{algorithm}

\subsection{Algorithm}
Algorithm \ref{alg:1} demonstrates the process of our meta-transfer training procedures of Section \ref{sec1} and \ref{sec2}. Lines 3-7 is the large-scale training stage. Lines 11-14 is the inner loop of meta-transfer learning where the base-learner is updated to task-specific loss. Lines 15-16 presents the meta-learner optimization.

Algorithm \ref{alg:2} presents the meta-test step, which is the zero-shot super-resolution. A few gradient updates ($n$) are performed while meta-test, and the super-resolved image is obtained with final updated parameters.

\section{Experiments}

\subsection{Training Details}
For the CNN, we adopt a simple $8$-layer CNN architecture with residual learning following ZSSR \cite{ZSSR}. Its number of parameters is $225$ K. For meta-transfer training, we use DIV2K \cite{DIV2K} for the high-quality dataset and we set $\alpha=0.01$ and $\beta=0.0001$ for entire training. For the inner loop, we conducted $5$ gradient updates, \ie $5$ unrolling steps, to obtain adapted parameters. We extracted training patches with a size of $64\times 64$. To cope with gradient vanishing or exploding problems due to the unrolling process of base learners, we utilize the weighted sum of losses from each step, \ie, providing supervision of additional losses to each unrolling step \cite{MAML++}. At the initial point, we evenly weigh the losses and decayed the weights except for the last unrolling step. In the end, the weighted loss converges to our final training task loss. We employ ADAM \cite{ADAM} optimizer as our meta-optimizer. As the subsampling process ($\downarrow_s$) can be the \emph{direct} method \cite{ZSSR} or the \emph{bicubic} subsampling \cite{SRMD, BlindSR}, we trained two models for different subsampling methods: \emph{direct} and \emph{bicubic}.

\subsection{Evaluations on ``Bicubic'' Downsampling}
\begin{table*}
	\begin{center}
			\begin{tabular}{|c|c|c|c||c|c|c|}
				\hline
				&\multicolumn{3}{c||}{Supervised}&\multicolumn{3}{c|}{Unsupervised}\\
				\hline\hline
				\rule[-1ex]{0pt}{3.5ex}
				Dataset & Bicubic & CARN \cite{CARN} & RCAN \cite{RCAN} & ZSSR \cite{ZSSR} & MZSR (1) & MZSR (10) \\
				\hline\hline
				\rule[-1ex]{0pt}{3.5ex}
				Set5 & 33.64/0.9293& 37.76/0.9590 & 38.18/0.9604&36.93/0.9554 &36.77/0.9549 &37.25/0.9567 \\
				BSD100 & 29.55/0.8427&  32.09/0.8978& 32.38/0.9018&31.43/0.8901 &31.33/0.8910 &31.64/0.8928 \\
				Urban100 &26.87/0.8398 &  31.92/0.9256& 33.30/0.9376&29.34/0.8941 &30.01/0.9054 & 30.41/0.9092\\
				\hline
			\end{tabular}
	\end{center}
	\caption{The average PSNR/SSIM results on ``bicubic'' downsampling scenario with $\times 2$ on benchmarks. The numbers in parenthesis in our methods stand for the number of gradient updates.}
	\label{table:bicubic}
\end{table*}

We evaluate our method with several recent SotA SISR methods, including supervised and unsupervised methods on famous benchmarks: Set5 \cite{Set5}, BSD100 \cite{B100}, and Urban100 \cite{Self-Exemplar}. We measure PSNR and SSIM \cite{SSIM} in Y-channel of YCbCr colorspace.

The overall results are shown in \tablename{~\ref{table:bicubic}}. CARN \cite{CARN} and RCAN \cite{RCAN}, which are trained for ``bicubic'' downsampling condition, show extremely overwhelming performances. Since the training scenario and the test scenario exactly match each other, supervision on external samples could boost the performance of CNN. On the other hands, ZSSR \cite{ZSSR} and our methods show improvements against bicubic interpolation but not as good as the supervised ones, because both methods are trained within the unsupervised or self-supervised regime. Our methods show comparable results to ZSSR within only \emph{one} single gradient descent update.

\subsection{Evaluations on Various Blur Kernels}
\begin{table*}
	\begin{center}
		\resizebox{0.9\linewidth}{!}{
			\begin{tabular}{|c|c|c|c|c||c|c|c|}
				\hline
				&\multicolumn{4}{c||}{Supervised}&\multicolumn{3}{c|}{Unsupervised}\\
				\hline\hline
				\rule[-1ex]{0pt}{3.5ex}
				Kernel & Dataset & Bicubic & RCAN \cite{RCAN} & IKC \cite{BlindSR} & ZSSR \cite{ZSSR} & MZSR (1) & MZSR (10)  \\
				\hline\hline
				\rule[-1ex]{0pt}{3.5ex}
				\multirow{3}{*}{$g^{d}_{0.2}$}&Set5&30.24/0.8976&  28.40/0.8618&29.09/0.8786&\textcolor{red}{34.29}/\textcolor{red}{0.9373}
				&33.14/0.9277 &\textcolor{blue}{33.74}/\textcolor{blue}{0.9301}\\
				&BSD100  & 27.45/0.7992& 25.16/0.7602&26.23/0.7808&\textcolor{red}{29.35}/\textcolor{red}{0.8465}&28.74/0.8389&\textcolor{blue}{29.03}/\textcolor{blue}{0.8415}\\
				&Urban100&24.70/0.7958& 21.68/0.7323&23.66/0.7806&\textcolor{red}{28.13}/\textcolor{red}{0.8788}&26.24/0.8394&\textcolor{blue}{26.60}/\textcolor{blue}{0.8439}\\
				\hline\hline
				\rule[-1ex]{0pt}{3.5ex}
				\multirow{3}{*}{$g^{d}_{2.0}$}&Set5&28.73/0.8449&29.15/0.8601&29.05/0.8896&34.90/0.9397&\textcolor{blue}{35.20}/\textcolor{blue}{0.9398}&\textcolor{red}{36.05}/\textcolor{red}{0.9439}\\
				&BSD100  &26.51/0.7157&26.89/0.7394&27.46/0.8156&30.57/\textcolor{blue}{0.8712}&\textcolor{blue}{30.58}/0.8627&\textcolor{red}{31.09}/\textcolor{red}{0.8739}\\
				&Urban100&23.70/0.7109&24.14/0.7384&25.17/0.8169&27.86/0.8582&\textcolor{blue}{28.23}/\textcolor{blue}{0.8657}&\textcolor{red}{29.19}/\textcolor{red}{0.8838}\\
				\hline\hline
				\rule[-1ex]{0pt}{3.5ex}
				\multirow{3}{*}{$g^{d}_{ani}$}&Set5&28.15/0.8265&28.42/0.8379&28.74/0.8565&33.96/\textcolor{blue}{0.9307}&\textcolor{blue}{34.05}/0.9271&\textcolor{red}{34.78}/\textcolor{red}{0.9323}\\
				&BSD100  &26.00/0.6891&26.22/0.7062&26.44/0.7310&\textcolor{red}{29.72}/\textcolor{red}{0.8479}&28.82/0.8013&\textcolor{blue}{29.54}/\textcolor{blue}{0.8297}\\
				&Urban100&23.13/0.6796&23.35/0.6982&23.62/0.7239&\textcolor{blue}{27.03}/\textcolor{blue}{0.8335}&26.51/0.8126&\textcolor{red}{27.34}/\textcolor{red}{0.8369}\\
				\hline\hline
				
				\rule[-1ex]{0pt}{3.5ex}
				\multirow{3}{*}{$g^{b}_{1.3}$}&Set5&30.54/0.8773&31.54/0.8992&33.88/0.9357&\textcolor{blue}{35.24}/\textcolor{blue}{0.9434}&35.18/0.9430&\textcolor{red}{36.64}/\textcolor{red}{0.9498}\\
				&BSD100  &27.49/0.7546&28.27/0.7904&\textcolor{blue}{30.95}/\textcolor{red}{0.8860}&30.74/0.8743&29.02/0.8544&\textcolor{red}{31.25}/\textcolor{blue}{0.8818}\\
				&Urban100&24.74/0.7527&25.65/0.7946&\textcolor{blue}{29.47}/\textcolor{blue}{0.8956}&28.30/0.8693&28.27/0.8771&\textcolor{red}{29.83}/\textcolor{red}{0.8965}\\
				\hline
			\end{tabular}
		}
	\end{center}
	\caption{The average PSNR/SSIM results on various kernels with $\times 2$ on benchmarks. The numbers in parenthesis in our methods stand for the number of gradient updates. The best results are highlighted in \textcolor{red}{red} and the second best are in \textcolor{blue}{blue}.}
	\label{table:Various}
\end{table*}

In this section, we demonstrate the results on various blur kernel conditions. We assume four scenarios: severe aliasing, isotropic Gaussian, unisotropic Gaussian, and isotropic Gaussisan followed by \emph{bicubic} subsampling. Precisely, the methods are
\begin{itemize}
	\item $g^{d}_{0.2}$: isotropic Gaussian blur kernel with width $\lambda=0.2$ followed by \emph{direct} subsampling.
	\item $g^{d}_{2.0}$: isotropic Gaussian blur kernel with width $\lambda=2.0$ followed by \emph{direct} subsampling.
	\item $g^{d}_{ani}$: anisotropic Gaussian with widths $\lambda_1 = 4.0$ and $\lambda_2 = 1.0$ with $\Theta=-0.5$ from Eq. \ref{eq:3}, followed by \emph{direct} subsampling.
	\item $g^{b}_{1.3}$: isotropic Gaussian blur kernel with width $\lambda=1.3$ followed by \emph{bicubic} subsampling.
	
\end{itemize} 

\begin{figure*}[t]
	\begin{center}	
		\begin{subfigure}[t]{0.28\linewidth}
			\centering
			\includegraphics[width=1\columnwidth]{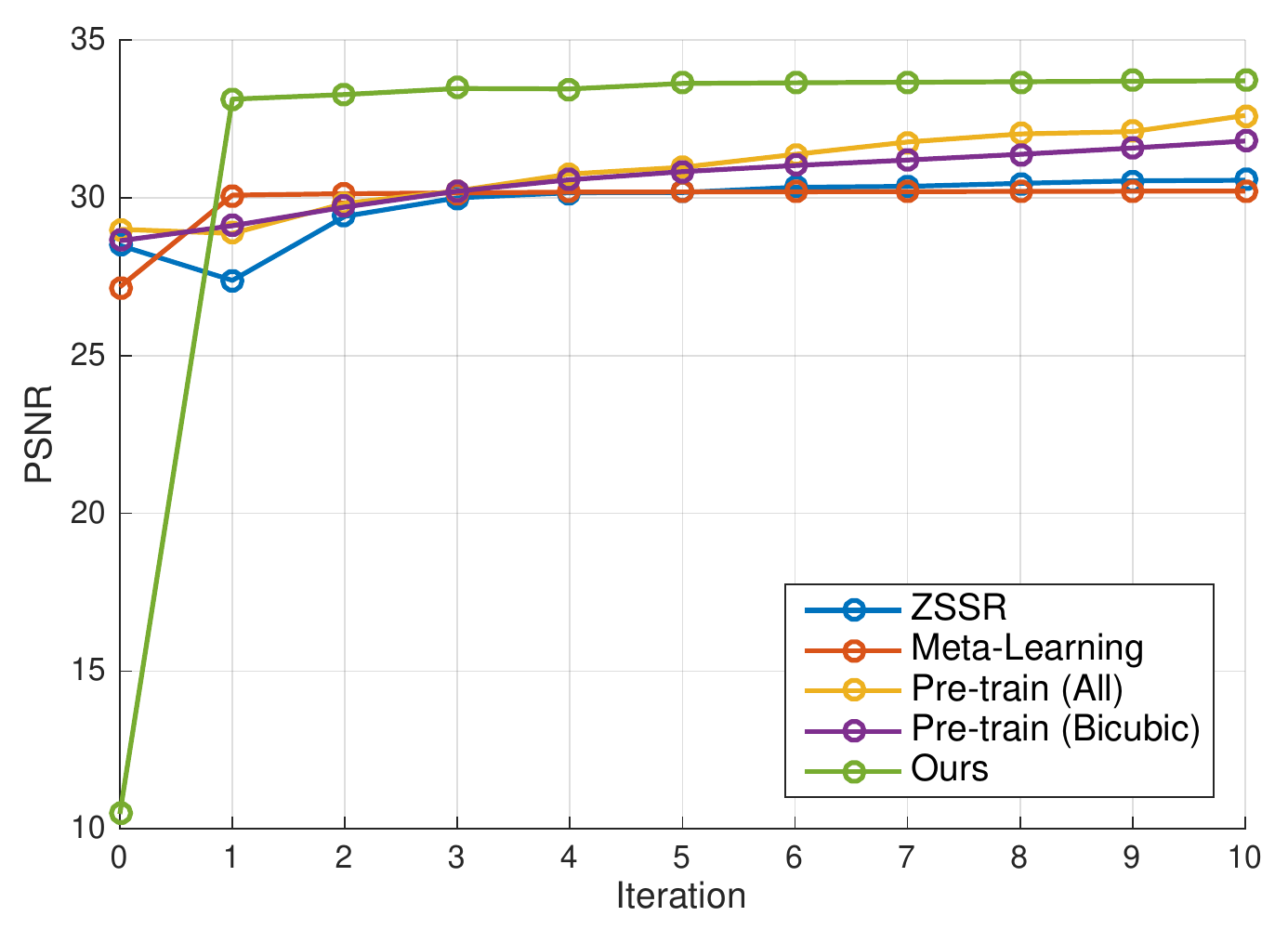}
			\caption{Aliased condition $g^{d}_{0.2}$}
		\end{subfigure}
		\begin{subfigure}[t]{0.28\linewidth}
			\centering
			\includegraphics[width=1\columnwidth]{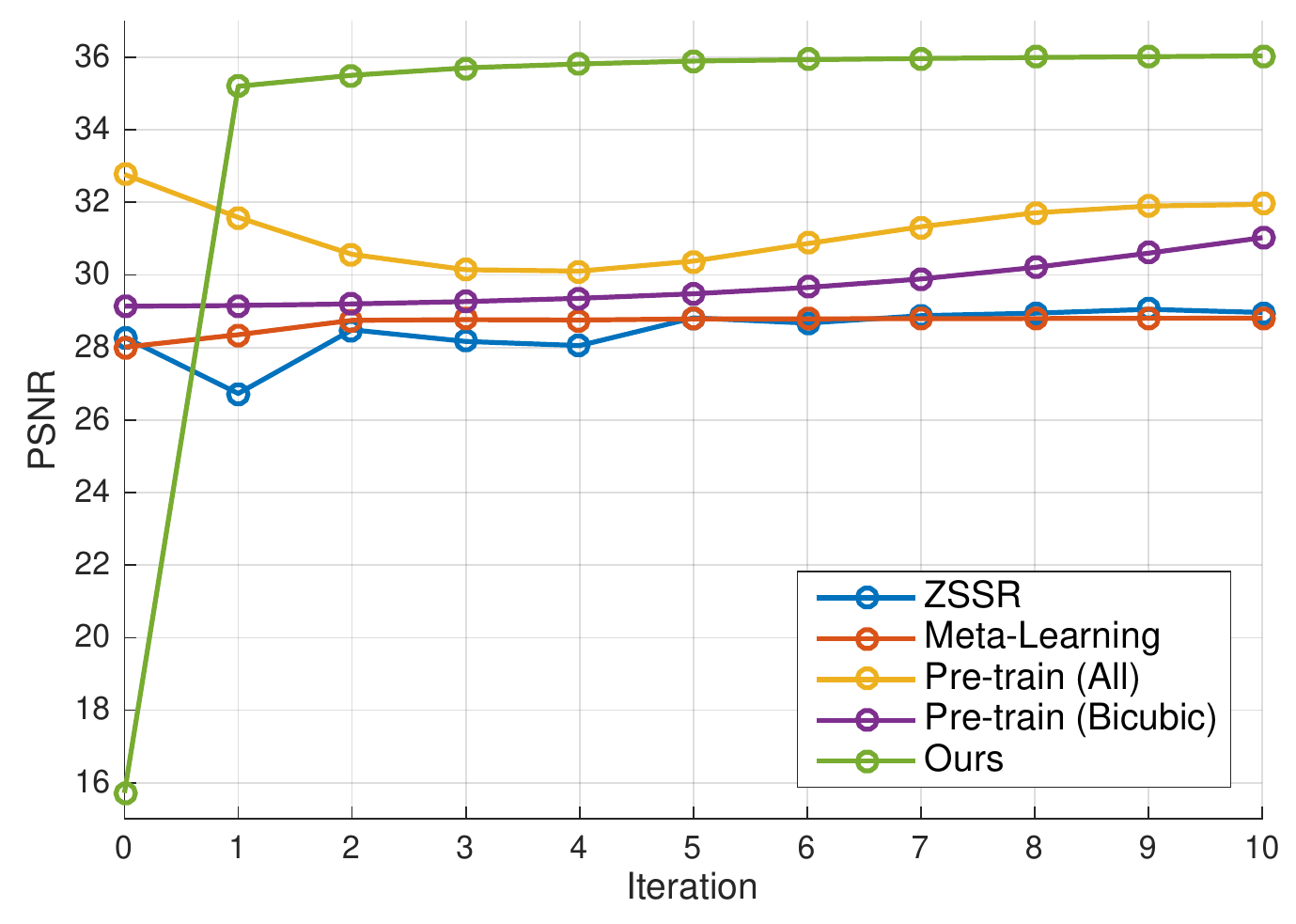}
			\caption{Isotropic Gaussian $g^{d}_{2.0}$  }
		\end{subfigure}
		\begin{subfigure}[t]{0.28\linewidth}
			\centering
			\includegraphics[width=1\columnwidth]{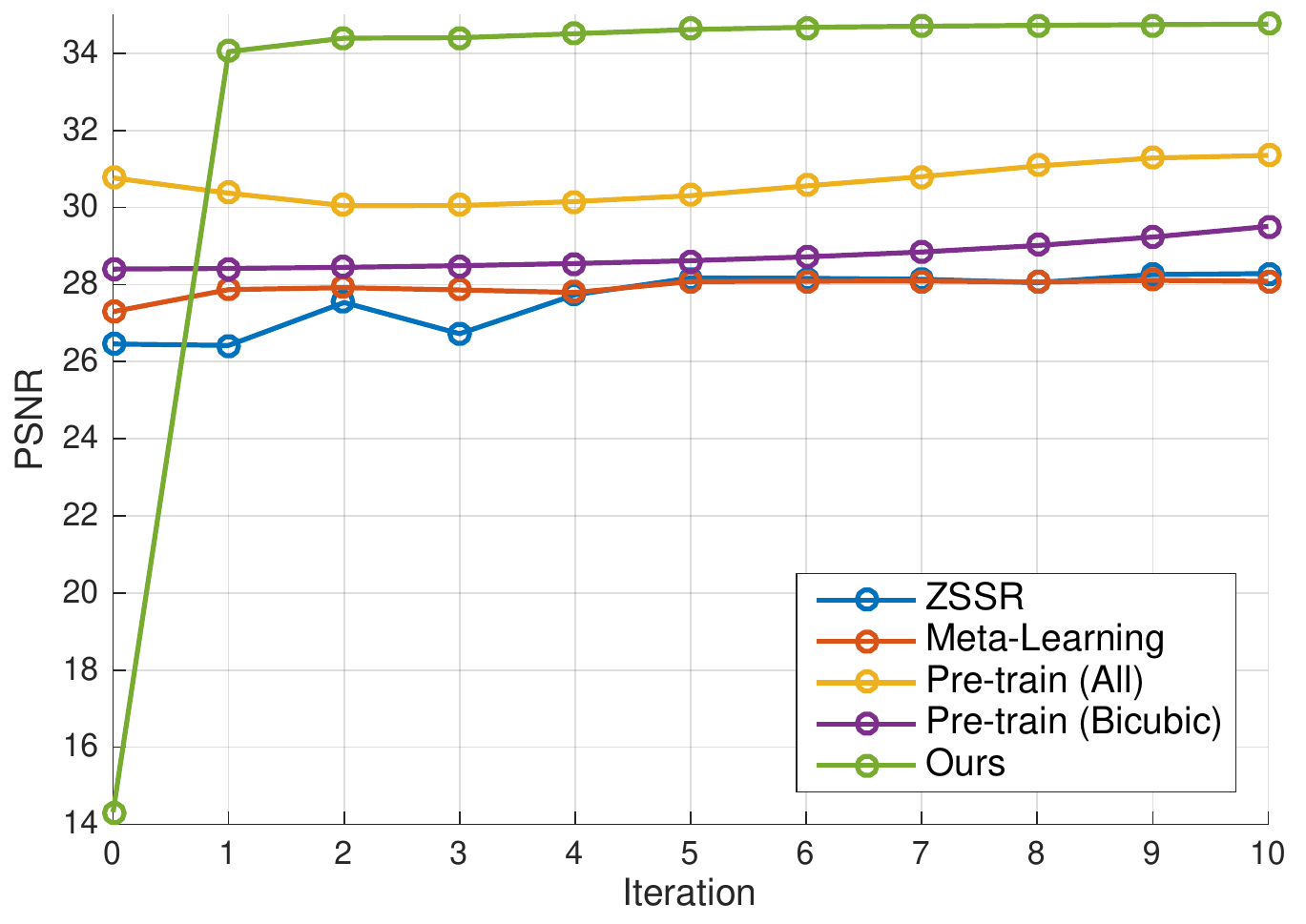}
			\caption{ Anisotropic Gaussian $g^{d}_{ani}$ }
		\end{subfigure}		
		
	\end{center}
	\caption{The average PSNR on Set5 vs. number of gradient update iterations. ``Meta-Learning'' is trained without initialization of pre-trained model. ``Pre-train (All)'' and ``Pre-train (Bicubic)'' are fine-tuned from pre-trained models for all kernels (blind model) and bicubic downsampling model, respectively. All methods except ours are optimized using ADAM \cite{ADAM} while our method is optimized with gradient descent.}
	\label{fig:iter}
\end{figure*}

\begin{figure*}[t]
	\captionsetup{justification=centering}
	\begin{center}	
		\begin{minipage}[t]{0.17\linewidth}
			\centering
			\raisebox{-0.7\height}{\includegraphics[width=1\columnwidth]{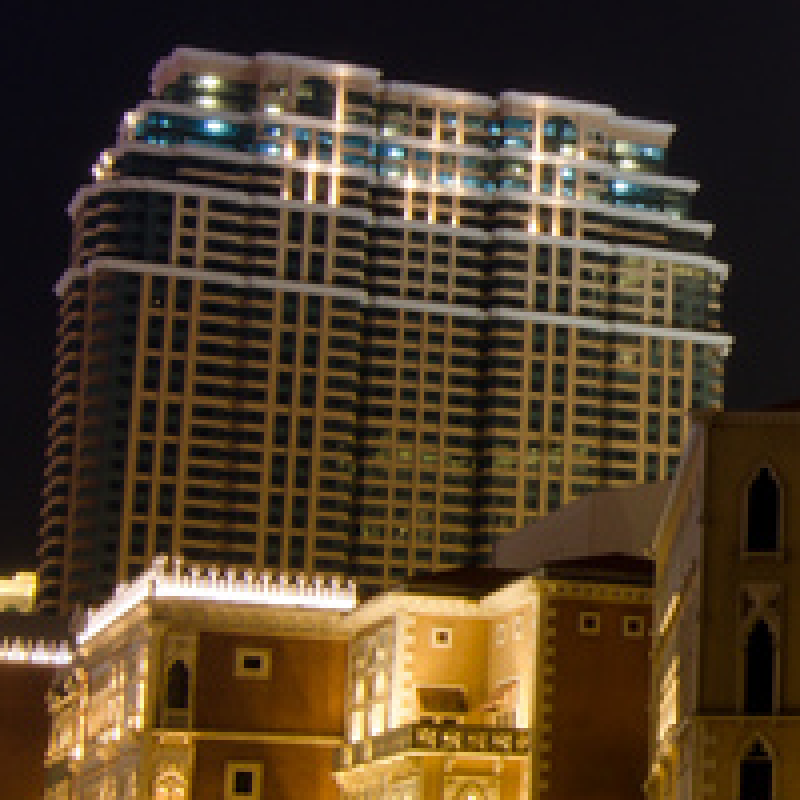}}
			\caption*{HR}
		\end{minipage}
	\quad
		\begin{minipage}[t]{0.78\linewidth}
		\begin{subfigure}[t]{0.235\linewidth}
			\centering
			\includegraphics[width=1\columnwidth]{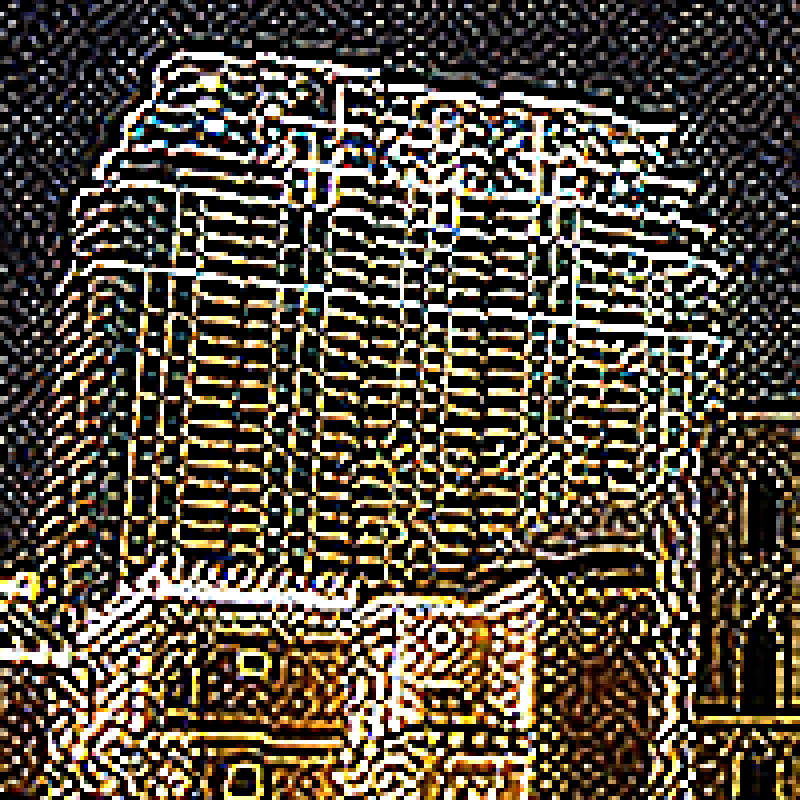}
			\caption{Initial point $g^{d}_{0.2}$ \\ $12.09$ dB}
		\end{subfigure}
		\begin{subfigure}[t]{0.235\linewidth}
			\centering
			\includegraphics[width=1\columnwidth]{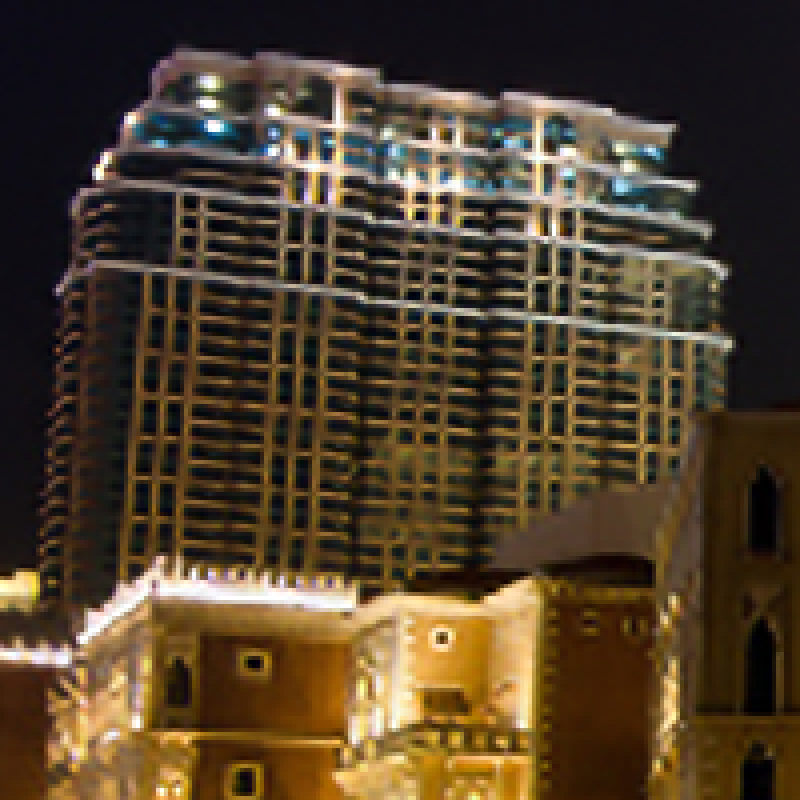}
			\caption{After one update $g^{d}_{0.2}$ \\ $30.27$ dB}
		\end{subfigure}
		\quad
		\begin{subfigure}[t]{0.235\linewidth}
			\centering
			\includegraphics[width=1\columnwidth]{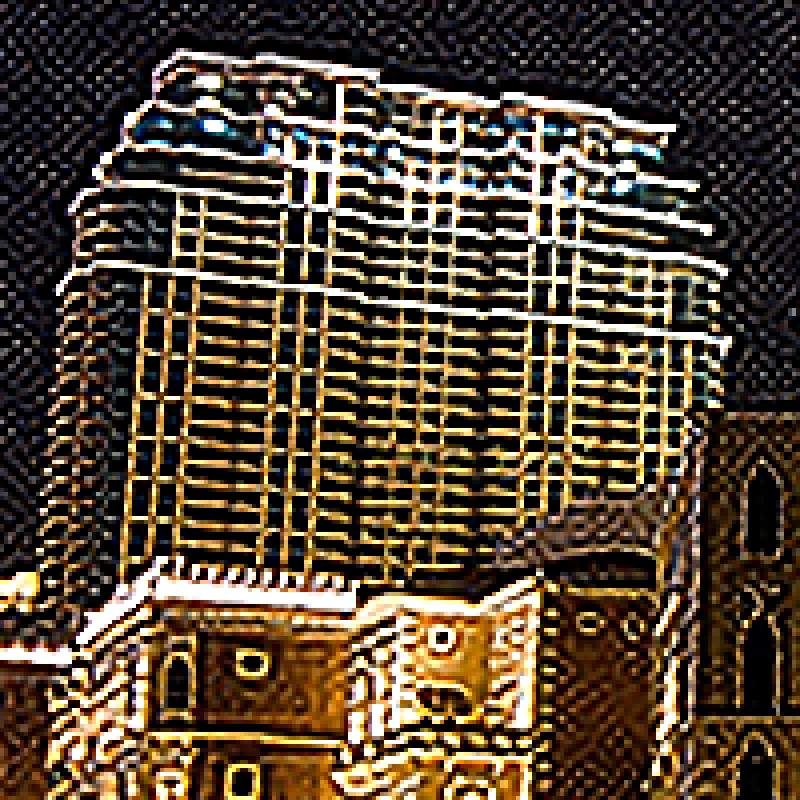}
			\caption{Initial point $g^{d}_{2.0}$ \\ $16.82$ dB}
		\end{subfigure}
		\begin{subfigure}[t]{0.235\linewidth}
			\centering
			\includegraphics[width=1\columnwidth]{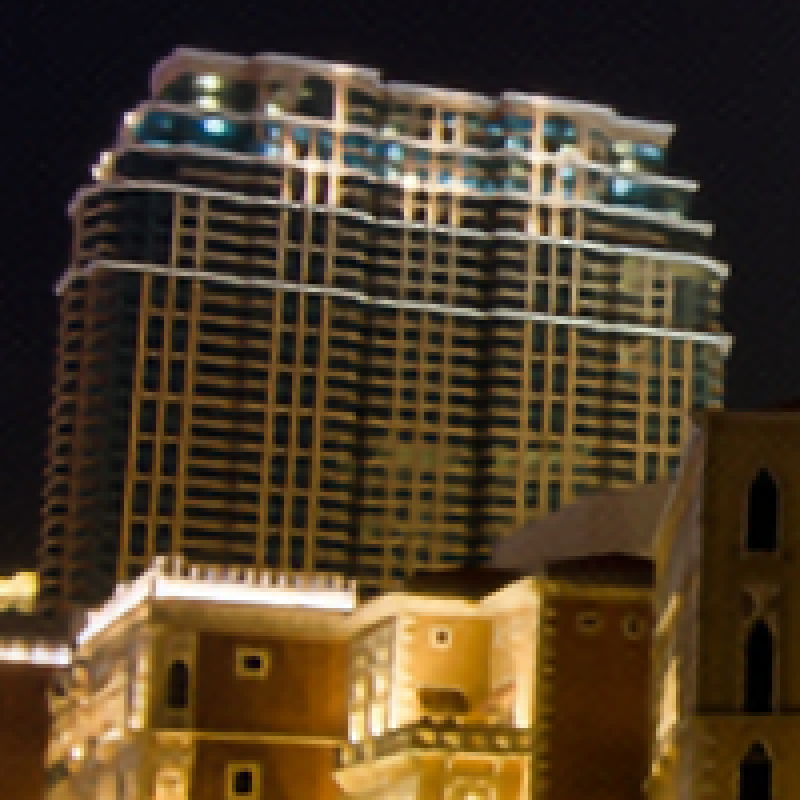}
			\caption{After one update $g^{d}_{2.0}$ \\ $33.08$ dB}
		\end{subfigure}
		\\
		\begin{subfigure}[t]{0.235\linewidth}
			\centering
			\includegraphics[width=1\columnwidth]{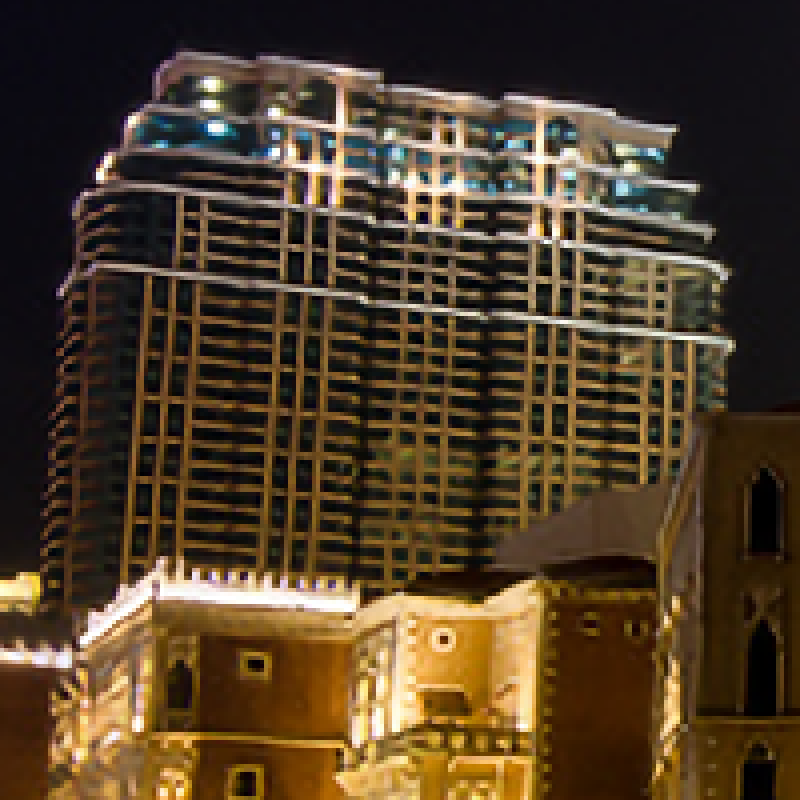}
			\caption{Initial point $g^{d}_{0.2}$ \\ $26.49$ dB}
		\end{subfigure}
		\begin{subfigure}[t]{0.235\linewidth}
			\centering
			\includegraphics[width=1\columnwidth]{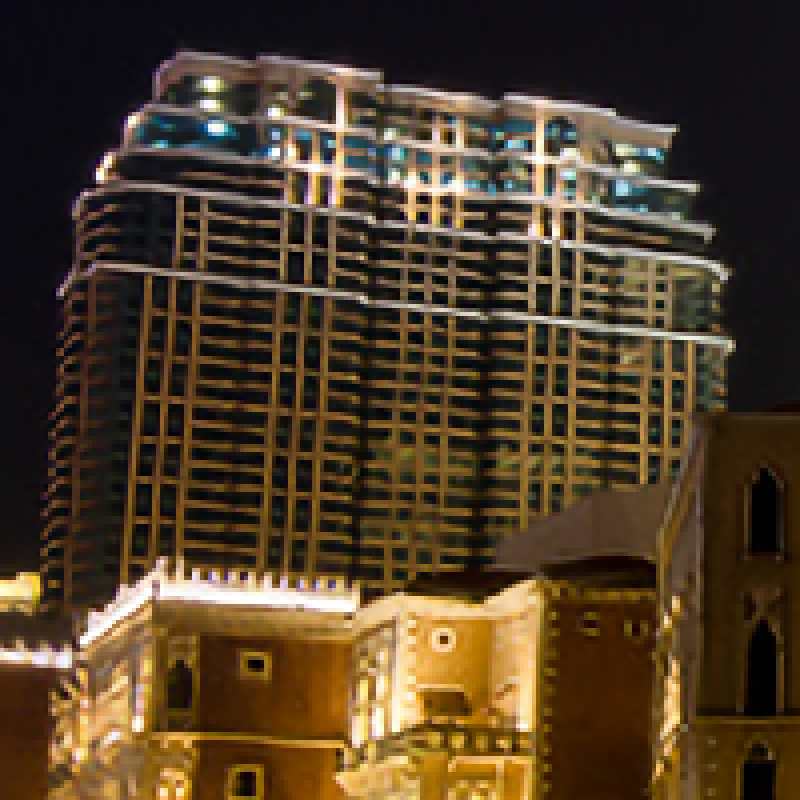}
			\caption{After one update $g^{d}_{0.2}$  \\ $26.87$ dB}
		\end{subfigure}
		\quad
		\begin{subfigure}[t]{0.235\linewidth}
			\centering
			\includegraphics[width=1\columnwidth]{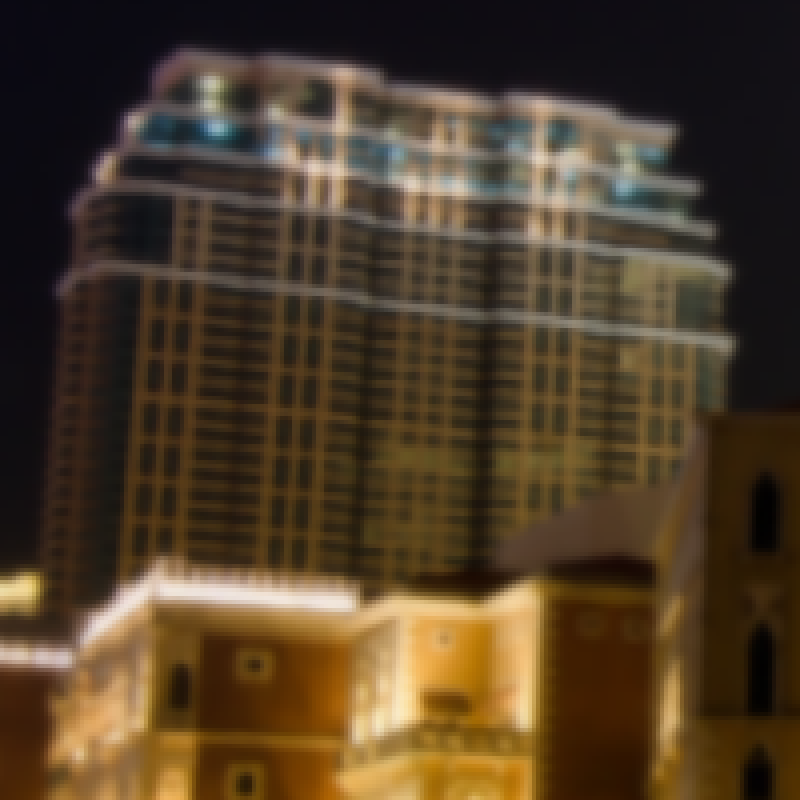}
			\caption{Initial point $g^{d}_{2.0}$ \\ $27.43$ dB}
		\end{subfigure}
		\begin{subfigure}[t]{0.235\linewidth}
			\centering
			\includegraphics[width=1\columnwidth]{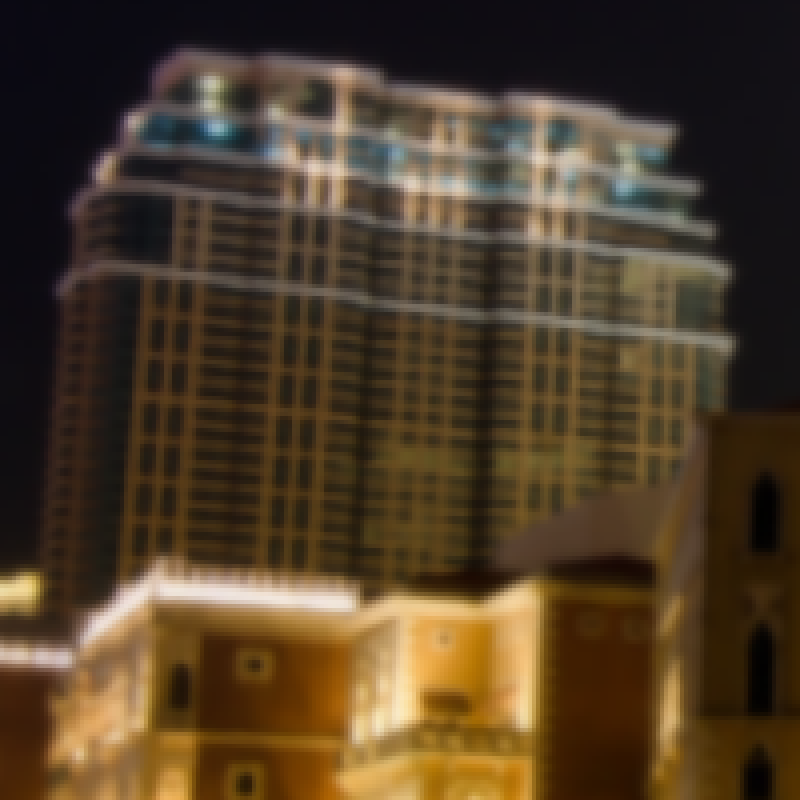}
			\caption{After one update $g^{d}_{2.0}$ \\ $27.47$ dB}
		\end{subfigure}
	\end{minipage}
		
	\end{center}
	
	\captionsetup{justification=raggedright,singlelinecheck=false}
	\caption{Visualization of the initial point and after one iteration of each method. Upper row images are from MZSR, and lower ones are from the pre-trained network on ``bicubic'' degradation.}
	\label{fig:002}
\end{figure*}

The results are shown in \tablename{~\ref{table:Various}}. As the SotA method RCAN \cite{RCAN} is trained on ``bicubic'' scenario, it shows inferior performance due to domain discrepancy and lack of flexibility.

For the case of aliasing ($g^d_{0.2}$), RCAN results are even worse than a simple bicubic interpolation method due to inconsistency between training and test condition. IKC\footnote{We reimplemented the code and retrained with DIV2K dataset.} \cite{BlindSR} is trained for
\emph{bicubic} subsampling, it never sees aliased images during training. Thus, it also shows a severe performance drop. On the other hand, ZSSR\footnote{We used the official code but without gradual configuration.} \cite{ZSSR} shows quite improved results due to its flexibility. However, it requires thousands of gradient updates, which require a large amount of time. Also, it starts from a random initial point and thus does not guarantee the same results for multiple tests. As shown in \tablename{~\ref{table:Various}}, our methods are comparable to others even with one single gradient update. Interestingly, our MZSR never sees the kernel with $\lambda=0.2$, but the CNN quickly adapts to specific image condition. In other words, compared to other methods, our method is more robust to extrapolation.

For other cases, which are isotropic and anisotropic Gaussian, our methods outperform others with a significantly large gap. In these cases, other methods have performance gains compared to bicubic interpolation, but the differences are minor. Similar tendencies of aliasing cases can be found in all other scenarios.
Interestingly, RCAN \cite{RCAN} shows slightly improved results compared to bicubic interpolation. Also, as the condition between training and test is consistent, IKC \cite{BlindSR} shows comparable results. Our methods also show remarkable performance in the case of \emph{bicubic} subsampling condition.
From the extensive experimental results, we believe that our MZSR is a fast, flexible, and accurate method for super-resolution.

\subsection{Real Image Super-Resolution}
To show the effectiveness of the proposed MZSR, we also conduct experiments on real images. Since there are no ground-truth images for real images, we only present the visual comparisons. Due to the page limit, all the comparisons on real images are presented in \emph{supplementary material}.

\section{Discussion}

\subsection{Number of Gradient Updates}

For ablation investigation, we train several models with different configurations. We assess the average PSNR results on Set5, which are shown in \figurename{~\ref{fig:iter}}. Interestingly, the initial point of our method shows the worst performance, but in one iteration, our method quickly adapts to the image condition and shows the best performance among the compared methods. Other methods sometimes show a slow increase in performance. In other words, they are not as flexible as ours in adapting to new image conditions.

We visualized the result at the initial point and after one gradient update in \figurename{~\ref{fig:002}}. As shown, the result of the initial point of MZSR is weird, but within one iteration, it is highly improved. On the other hand, the result of a pre-trained network is more natural than MZSR, but its improvement after one gradient update is minor.
Furthermore, it is shown that the performance of our method increases as the gradient descent update progresses, despite the fact that it is trained for maximum performance after five gradient steps. This result suggests that with more gradient update iterations, we might expect more of the performance improvements.

\subsection{Multi-scale Models}

\begin{table}
	\begin{center}
				\resizebox{\linewidth}{!}{
		\begin{tabular}{|c|c|c|c|}
			\hline
			\rule[-1ex]{0pt}{3.5ex}
			 PSNR (dB) &$g^{d}_{0.2}$ &$g^{d}_{2.0}$ &$g^{d}_{ani}$  \\
			\hline\hline
			\rule[-1ex]{0pt}{3.5ex}
			Multi-scale (10) &$33.33 (-0.41)$& $35.67 (-0.97)$ &$33.95 (-0.83)$ \\
			\hline
		\end{tabular}
				}
	\end{center}
	\caption{Average PSNR results of multi-scale model on Set5 with $\times 2$. The number in parenthesis is PSNR loss compared to the single-scale model.}
	\label{table:multi}
\end{table}

\begin{figure}[t]
	\begin{center}	
		\begin{subfigure}[t]{0.4\linewidth}
			\centering
			\includegraphics[width=1\columnwidth]{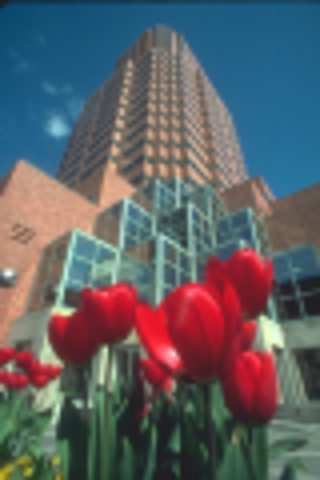}
			\caption{Bicubic interpolation}
		\end{subfigure}
		\begin{subfigure}[t]{0.4\linewidth}
			\centering
			\includegraphics[width=1\columnwidth]{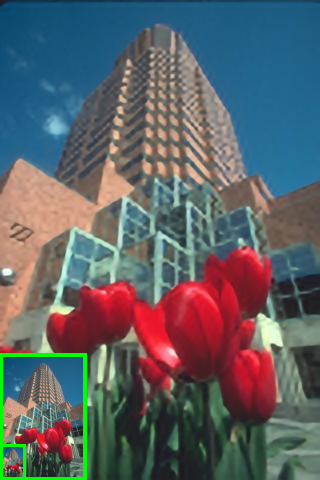}
			\caption{MZSR (Ours)}
		\end{subfigure}		
	\end{center}
	\caption{MZSR results on scaling factor $\times 4$ with blur kernel $g^{d}_{2.0}$. Despite the size of LR son image is $30\times20$, MZSR learns internal information. (Green boxes at the lower left corner of MZSR image are $\mathbf{I}_{son}$ and $\mathbf{I}_{LR}$)}
	\label{fig:multi}
\end{figure}

We additionally trained a multi-scale model with the scaling factors $s\in[2.0,4.0]$. The results on $\times 2$ show worse results comparable to single-scale model as shown in \tablename{~\ref{table:multi}}. With multiple scaling factors, the task distribution $p(\mathcal{T})$ becomes more complex, in which the meta-learner struggles to capture such regions that are suitable for fast adaptation.

Moreover, when meta-testing larger scaling factors, the size of $\mathbf{I}_{son}$ becomes too small to provide enough information to the CNN. Hence, the CNN rarely utilizes information from a very small LR son image.
Importantly, as our CNN learns internal information of CNN, such images with multi-scale recurrent patterns show plausible results even with large scaling factors, as shown in \figurename{~\ref{fig:multi}}.

\subsection{Complexity}
\begin{table}
	\begin{center}
			\begin{tabular}{|c|r|r|}
				\hline
				\rule[-1ex]{0pt}{3.5ex}
				Methods & Parameters & Time (sec) \\
				\hline\hline
				\rule[-1ex]{0pt}{3.5ex}
				CARN \cite{CARN}& 1,592 K & 0.47 \\
								
				\rule[-1ex]{0pt}{3.5ex}
				RCAN \cite{RCAN}& 15,445 K & 1.72 \\
				\hline\hline				
				\rule[-1ex]{0pt}{3.5ex}
				ZSSR \cite{ZSSR}& 225 K & 142.72 \\
								
				\rule[-1ex]{0pt}{3.5ex}
				MZSR (1) & 225 K & 0.13 \\
				
				\rule[-1ex]{0pt}{3.5ex}
				MZSR (10) & 225 K & 0.36 \\		
				\hline
			\end{tabular}
	\end{center}
	\caption{Comparisons of the number of parameters and time complexity for super-resolution of $256\times 256$ LR image with scaling factor $\times 2$.}
	\label{table:complexity}
\end{table}

We evaluate the overall model and time complexities for several comparisons, and the results are shown in \tablename{~\ref{table:complexity}}.
We measure time on the environment of NVIDIA Titan XP GPU.
Two fully-supervised feedforward networks for ``bicubic'' degradation, CARN and RCAN, require a large number of parameters. Even though CARN is proposed as a lightweight network which requires one-tenth of parameters compared to RCAN, it still requires much more parameters compared to unsupervised networks. However, the time consumptions for both model are quite comparable, because only feedforward computation is involved.

On the other hand, ZSSR, which is totally unsupervised, requires much less number of parameters due to the image-specific CNN. However, it requires thousands of forward and backward pass to get a super-resolved image, \ie, a large amount of time exceeding a practical extent. Our method MZSR with a single gradient update requires the shortest time among comparisons. Also, even with $10$ iterations of the backward pass, our method still shows comparable time consumption against CARN.

\section{Conclusion}
In this paper, we have presented a fast, flexible, and lightweight self-supervised super-resolution method by exploiting both external and internal samples. Specifically, we adopt an optimization-based meta-learning method jointly with transfer learning to seek an initial point that is sensitive to different conditions of blur kernels. Therefore, our method can quickly adapt to specific image conditions within a few gradient updates. From our extensive experiments, we show that our MZSR outperforms other methods, including ZSSR, which requires thousands of gradient descent iterations. Furthermore, we demonstrate the effectiveness of our method with complexity evaluation. Yet, there are lots of parts that can be improved from our work such as network architecture, learning strategies, and multi-scale model, and we leave these as future works. Our code is publicly available at \url{https://www.github.com/JWSoh/MZSR}.

\paragraph{Acknowledgements}
This research was supported in part by Projects for Research and Development of Police science and Technology under Center for Research and Development of Police science and Technology and Korean National Police Agency (PA-C000001), and in part by Samsung Electronics Co., Ltd.

\clearpage

\section*{Appendix}

\section*{A. Evaluation on Scaling Factor $\times4$}
\begin{table*}[t]
	\begin{center}
		\resizebox{\linewidth}{!}{
			\begin{tabular}{|c|c|c|c|c||c|c|c|}
				\hline
				$g^{d}_{2.0}$&\multicolumn{3}{c||}{Supervised}&\multicolumn{4}{c|}{Unsupervised}\\
				\hline\hline
				\rule[-1ex]{0pt}{3.5ex}
				Dataset & Bicubic & RCAN \cite{RCAN} & IKC \cite{BlindSR} & ZSSR \cite{ZSSR} & Multi-scale (1) & MZSR (1) & MZSR (10) \\
				\hline\hline
				\rule[-1ex]{0pt}{3.5ex}
				Set5 & 24.74/0.7321&23.92/0.7283 &24.01/0.7322&27.39/0.7685&29.85/0.8601&\textcolor{blue}{30.20}/\textcolor{blue}{0.8655}&\textcolor{red}{30.50}/\textcolor{red}{0.8704}\\
				BSD100 &24.01/0.5998&23.16/0.5918&23.12/0.5939&25.89/0.6776&26.68/0.7136&\textcolor{blue}{26.73}/\textcolor{blue}{0.7138}&\textcolor{red}{26.89}/\textcolor{red}{0.7168}\\
				Urban100 &21.16/0.5811&19.52/0.5400&19.81/0.5583&23.53/0.6822&24.13/0.7251&\textcolor{blue}{24.36}/\textcolor{blue}{0.7333}&\textcolor{red}{24.65}/\textcolor{red}{0.7394}\\
				\hline
			\end{tabular}
		}
	\end{center}
	\caption{Average PSNR/SSIM results on the scaling factor $\times 4$ on benchmarks. The numbers in parenthesis in our methods stand for the number of gradient updates. The best and the second best are highlighted in \textcolor{red}{red} and \textcolor{blue}{blue}, respectively.}
	\label{table:result4}
\end{table*}

To evaluate the performance on large scaling factors, we demonstrate the results on scaling factor $\times4$ with isotropic Gaussian kernel with width $2.0$ in \tablename{~\ref{table:result4}}. As shown, our methods show comparable results to others even with one gradient update, for large scaling factors too. Also, we found that multi-scale model shows worse results than a single-scale model as evidenced in the scaling factor $\times2$.

\section*{B. Effects of Kernels on Meta-test Time}
\begin{figure*}[t]
	\begin{center}
		\captionsetup{justification=centering}
		\begin{subfigure}[t]{0.16\linewidth}
			\centering
			\includegraphics[width=1\columnwidth]{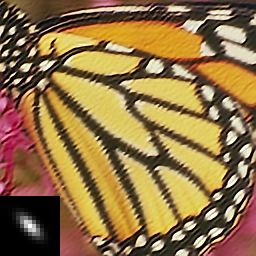}
			\caption{21.40 dB}
		\end{subfigure}
		\begin{subfigure}[t]{0.16\linewidth}
			\centering
			\includegraphics[width=1\columnwidth]{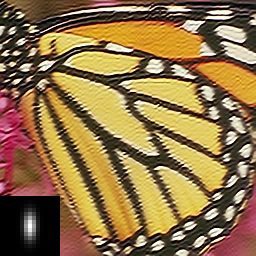}
			\caption{19.52 dB}
		\end{subfigure}
		\begin{subfigure}[t]{0.16\linewidth}
			\centering
			\includegraphics[width=1\columnwidth]{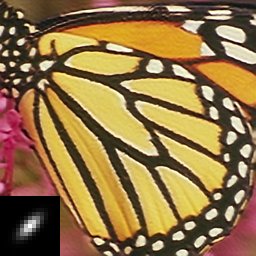}
			\caption{27.95 dB}
		\end{subfigure}
		\begin{subfigure}[t]{0.16\linewidth}
			\centering
			\includegraphics[width=1\columnwidth]{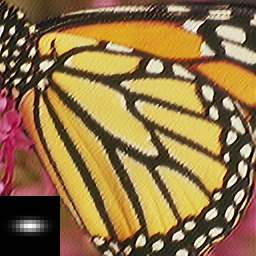}
			\caption{24.52 dB}
		\end{subfigure}
		\begin{subfigure}[t]{0.16\linewidth}
			\centering
			\includegraphics[width=1\columnwidth]{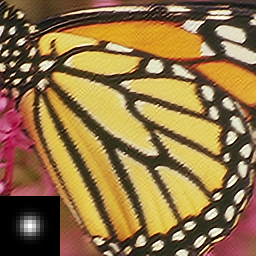}
			\caption{24.32 dB}
		\end{subfigure}
		\begin{subfigure}[t]{0.16\linewidth}
			\centering
			\includegraphics[width=1\columnwidth]{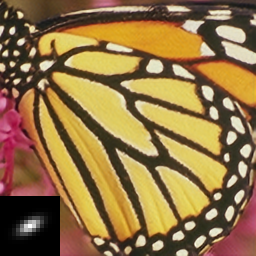}
			\caption{30.31 dB}
		\end{subfigure}
		
	\end{center}
	\caption{Comparisons when different kernels are applied on meta-test time. The last result is when the true kernel is applied.}
	\label{fig:mismatch}
\end{figure*}

To evaluate the effects of input kernels on meta-test time, we obtained several results by feeding various kernels. The results are shown in \figurename{~\ref{fig:mismatch}}. It is obvious that kernel mismatch degrades the output result severely. Especially, when the input kernel largely deviates from the true kernel, the result is not very pleasing as shown in \figurename{~\ref{fig:mismatch}}(a) and (b). However, if the input kernel has similar shape as the true kernel then the result looks quite plausible as shown in \figurename{~\ref{fig:mismatch}}(c). In conclusion, the kernel estimation or knowing the true kernel is crucial for the performance gain with our method.

\section*{C. Visualization}

\begin{figure*}[t]
	\begin{center}
		\captionsetup{justification=centering}
		
		\begin{subfigure}[t]{0.18\linewidth}
			\centering
			\includegraphics[width=1\columnwidth]{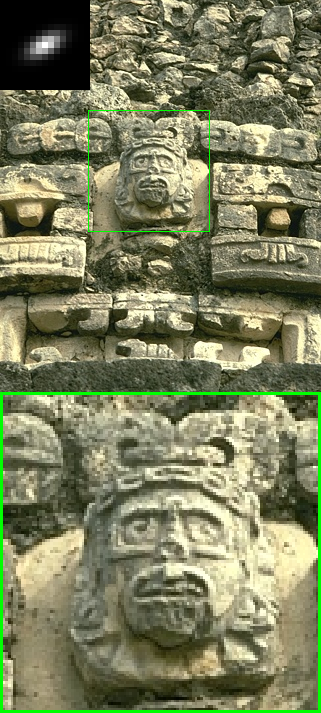}
			\caption*{GT}
		\end{subfigure}
		\begin{subfigure}[t]{0.18\linewidth}
			\centering
			\includegraphics[width=1\columnwidth]{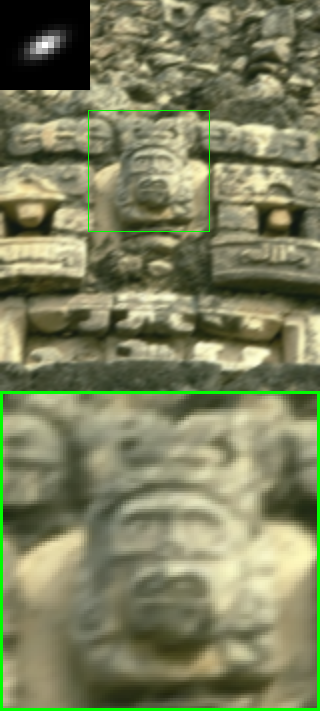}
			\caption*{Bicubic}
		\end{subfigure}
		\begin{subfigure}[t]{0.18\linewidth}
			\centering
			\includegraphics[width=1\columnwidth]{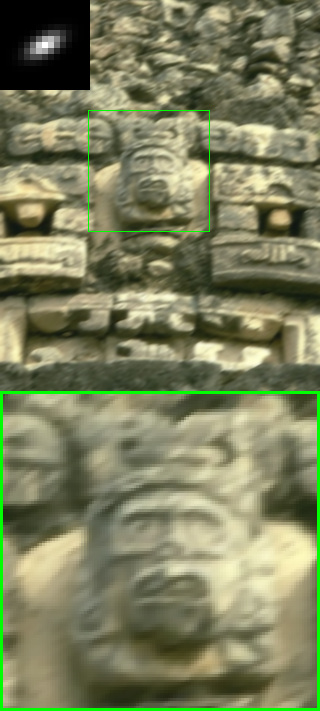}
			\caption*{RCAN \cite{RCAN}}
		\end{subfigure}
		\begin{subfigure}[t]{0.18\linewidth}
			\centering
			\includegraphics[width=1\columnwidth]{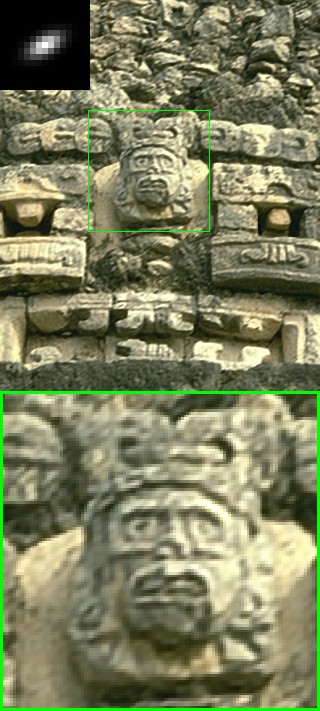}
			\caption*{ZSSR \cite{ZSSR} \\ \emph{2,160 updates}}
		\end{subfigure}
		\begin{subfigure}[t]{0.18\linewidth}
			\centering
			\includegraphics[width=1\columnwidth]{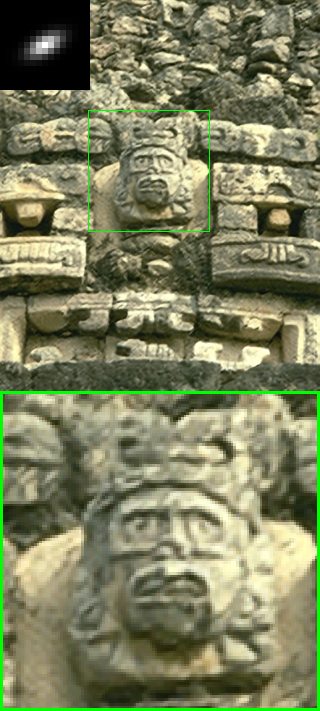}
			\caption*{MZSR (Ours) \\ \emph{10 updates}}
		\end{subfigure}
		
	\end{center}
	\caption{Visualized comparisons of super-resolution results ($\times 2$) with anisotropic blur kernel $g^{d}_{ani}$.}
	\label{fig:supp001}
\end{figure*}

\begin{figure*}[t]
	\begin{center}
		\captionsetup{justification=centering}
		
		\begin{subfigure}[t]{0.18\linewidth}
			\centering
			\includegraphics[width=1\columnwidth]{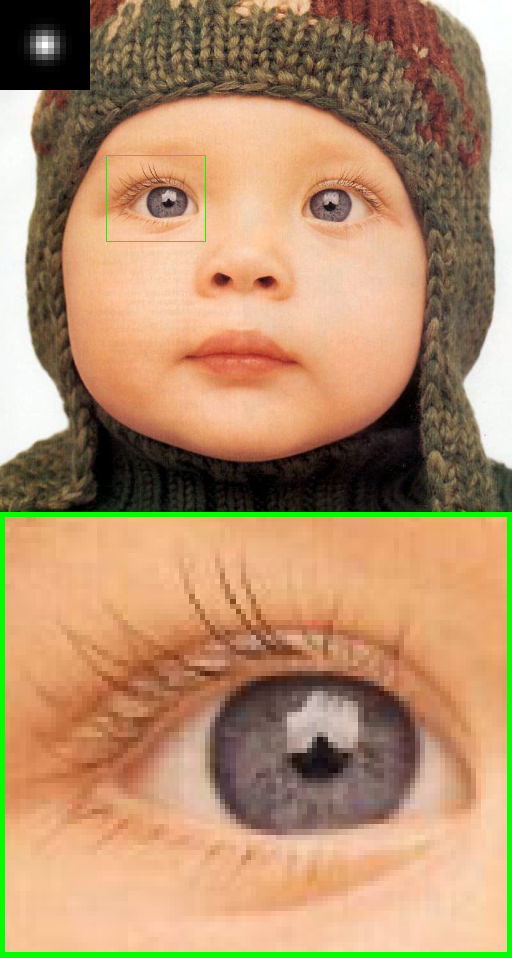}
			\caption*{GT}
		\end{subfigure}
		\begin{subfigure}[t]{0.18\linewidth}
			\centering
			\includegraphics[width=1\columnwidth]{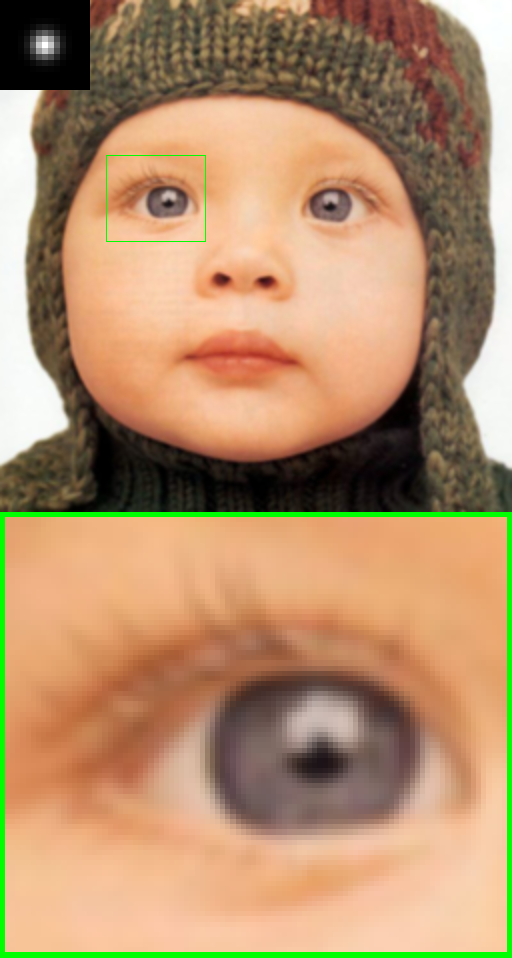}
			\caption*{Bicubic}
		\end{subfigure}
		\begin{subfigure}[t]{0.18\linewidth}
			\centering
			\includegraphics[width=1\columnwidth]{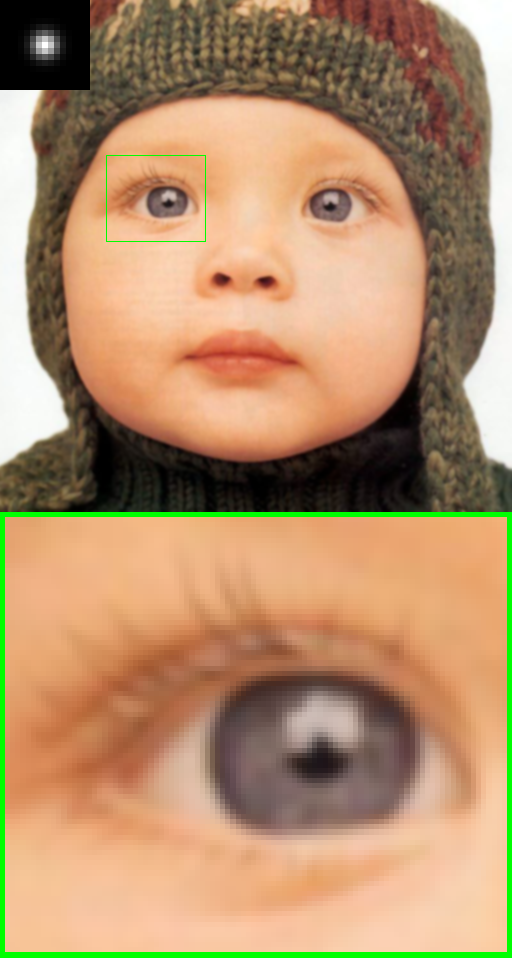}
			\caption*{RCAN \cite{RCAN}}
		\end{subfigure}
		\begin{subfigure}[t]{0.18\linewidth}
			\centering
			\includegraphics[width=1\columnwidth]{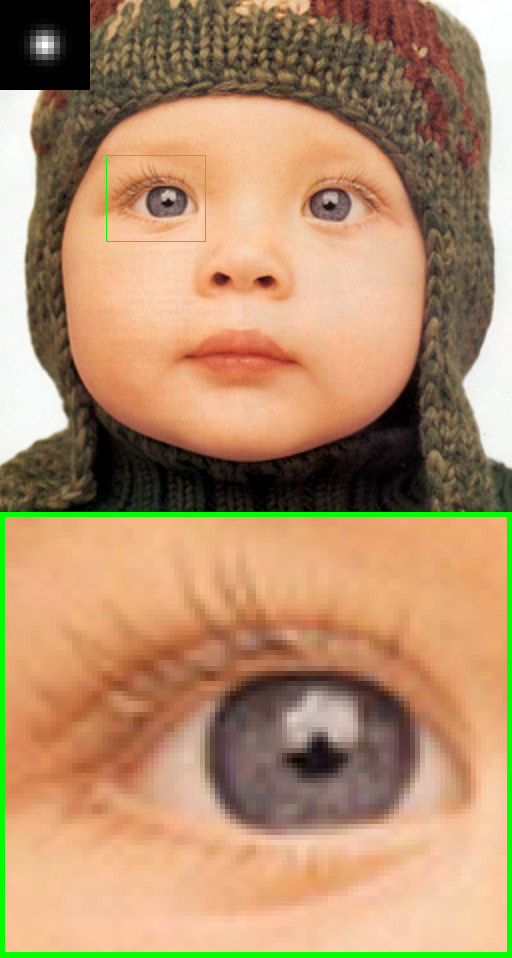}
			\caption*{ZSSR \cite{ZSSR} \\ \emph{2,600 updates}}
		\end{subfigure}
		\begin{subfigure}[t]{0.18\linewidth}
			\centering
			\includegraphics[width=1\columnwidth]{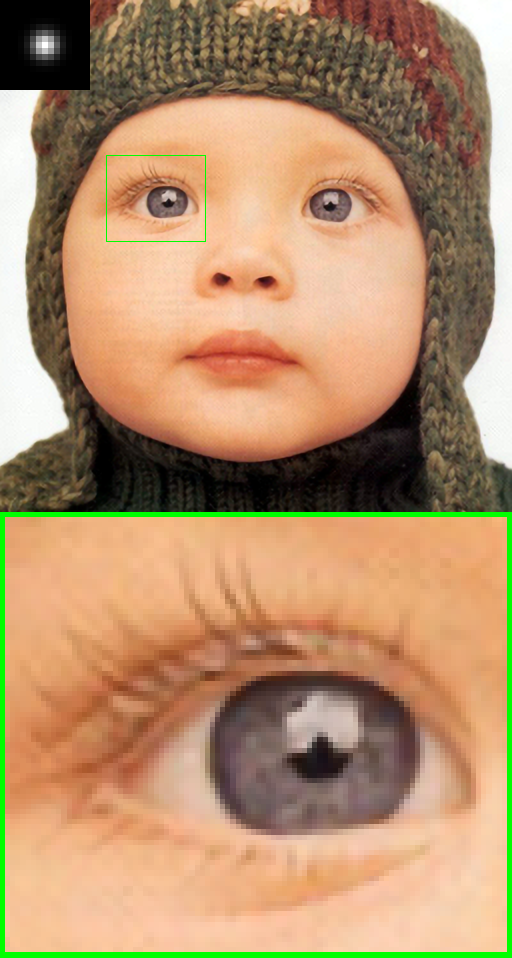}
			\caption*{MZSR (Ours) \\ \emph{10 updates}}
		\end{subfigure}
		
	\end{center}
	\caption{Visualized comparisons of super-resolution results ($\times 2$) with isotropic blur kernel and bicubic subsampling $g^{b}_{1.3}$.}
	\label{fig:supp003}
\end{figure*}

\begin{figure*}[t]
	\begin{center}
		\captionsetup{justification=centering}
		
		\begin{subfigure}[t]{0.24\linewidth}
			\centering
			\includegraphics[width=1\columnwidth]{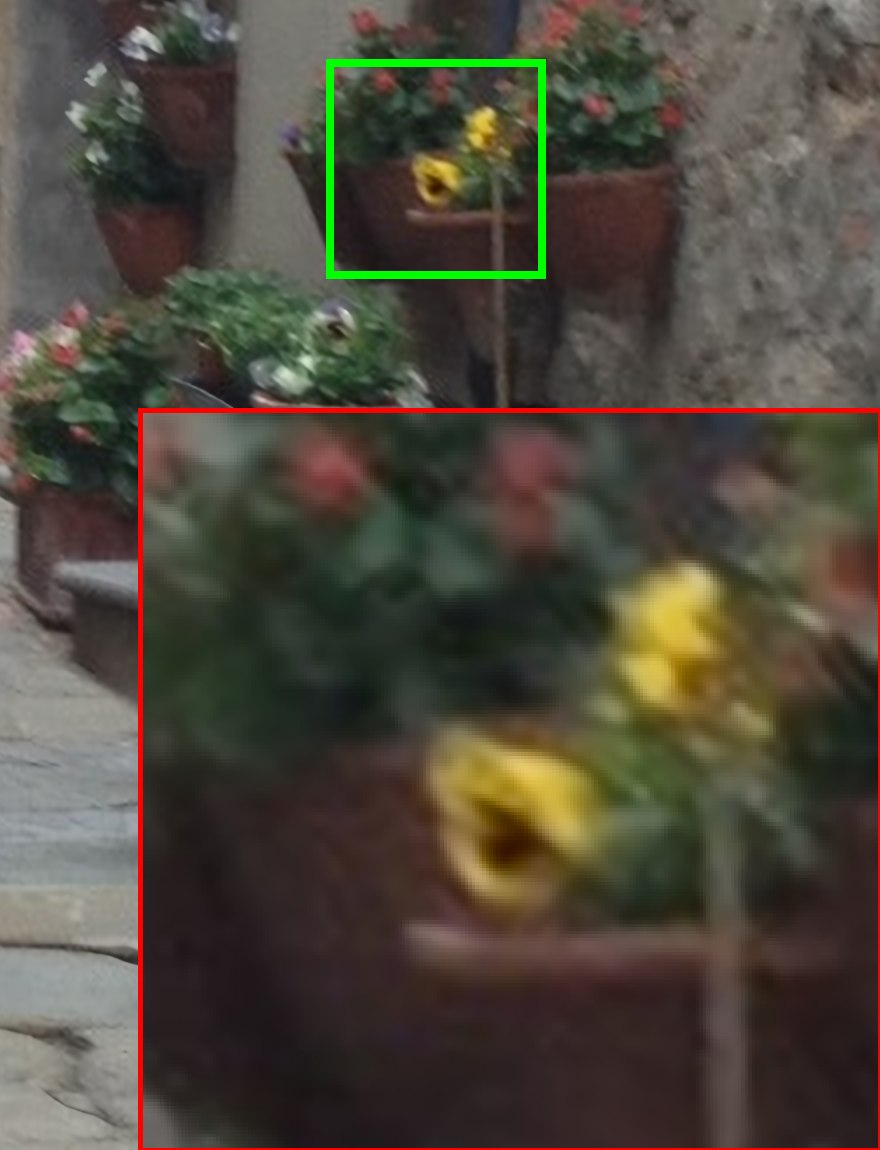}
			\caption*{CARN \cite{CARN}}
		\end{subfigure}
		\begin{subfigure}[t]{0.24\linewidth}
			\centering
			\includegraphics[width=1\columnwidth]{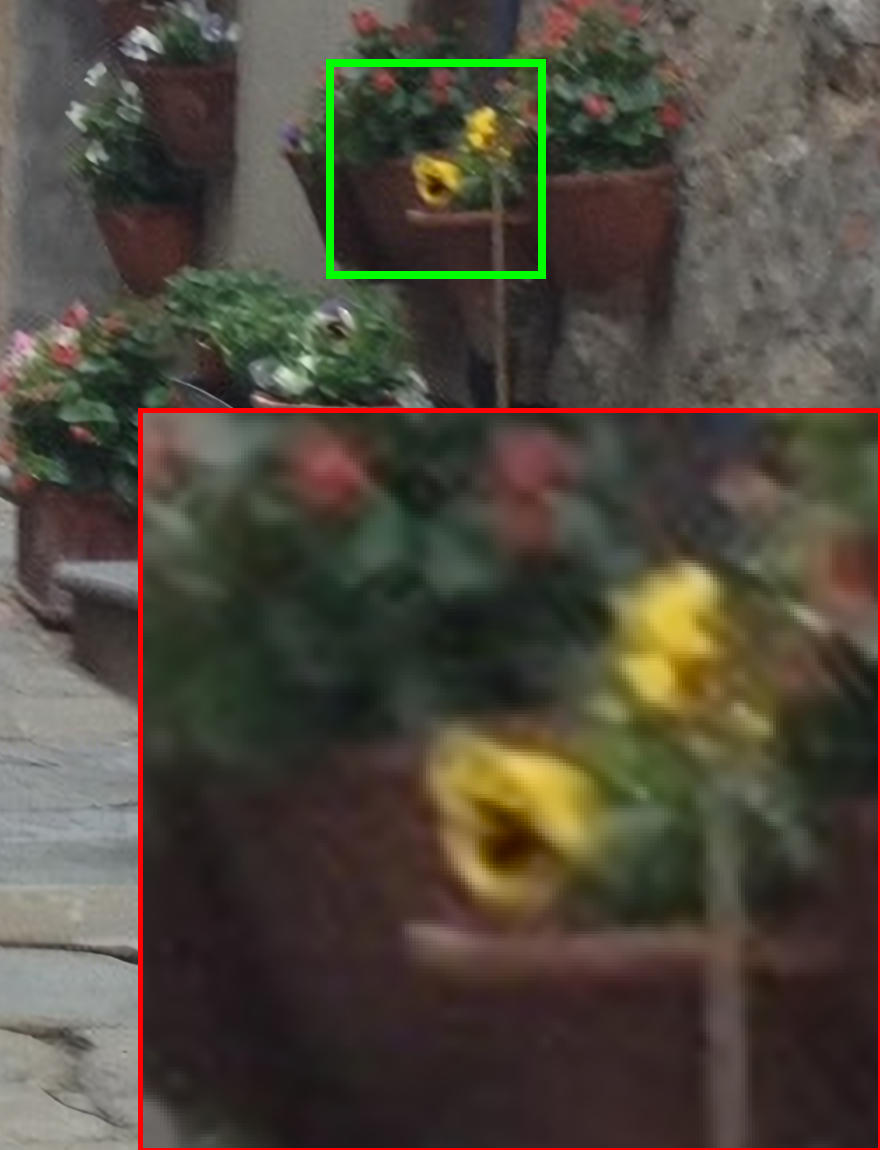}
			\caption*{IKC \cite{BlindSR}}
		\end{subfigure}
		\begin{subfigure}[t]{0.24\linewidth}
			\centering
			\includegraphics[width=1\columnwidth]{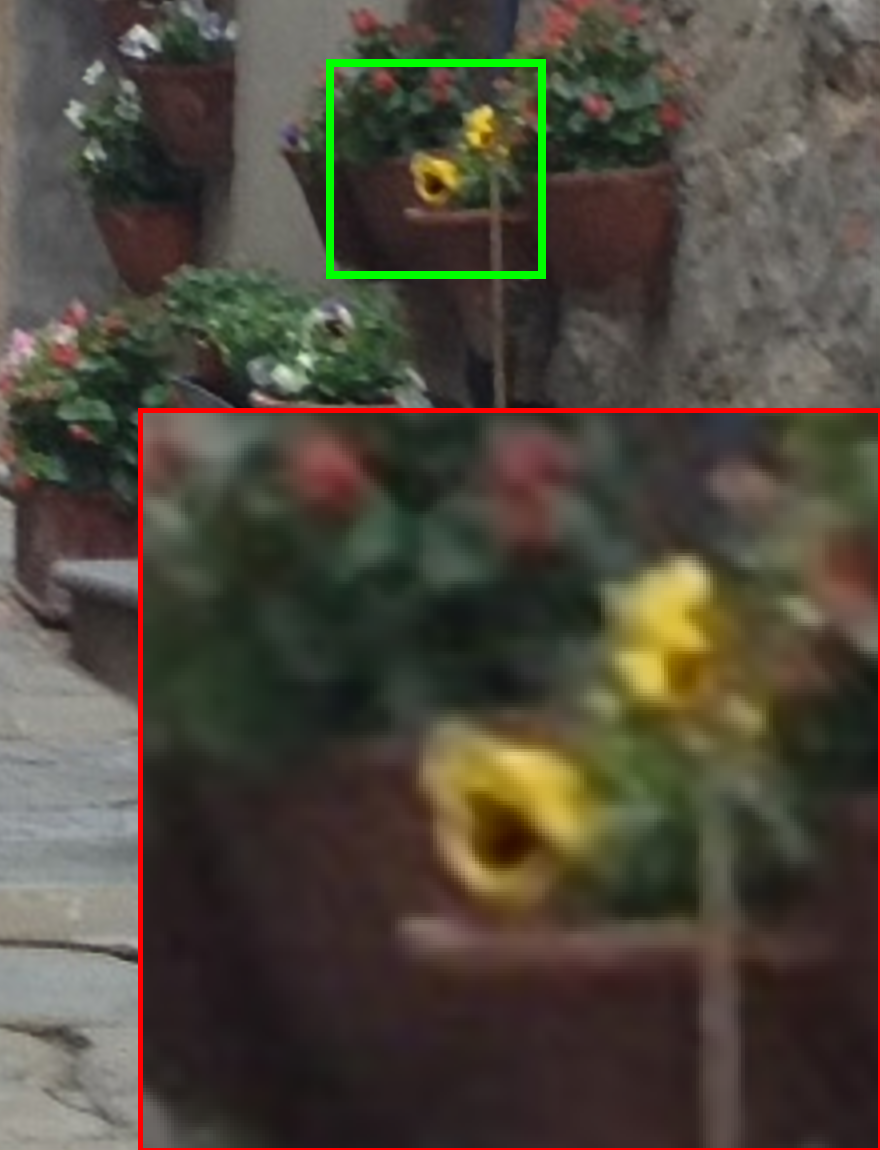}
			\caption*{ZSSR \cite{ZSSR} \\ \emph{1,500 updates}}
		\end{subfigure}
		\begin{subfigure}[t]{0.24\linewidth}
			\centering
			\includegraphics[width=1\columnwidth]{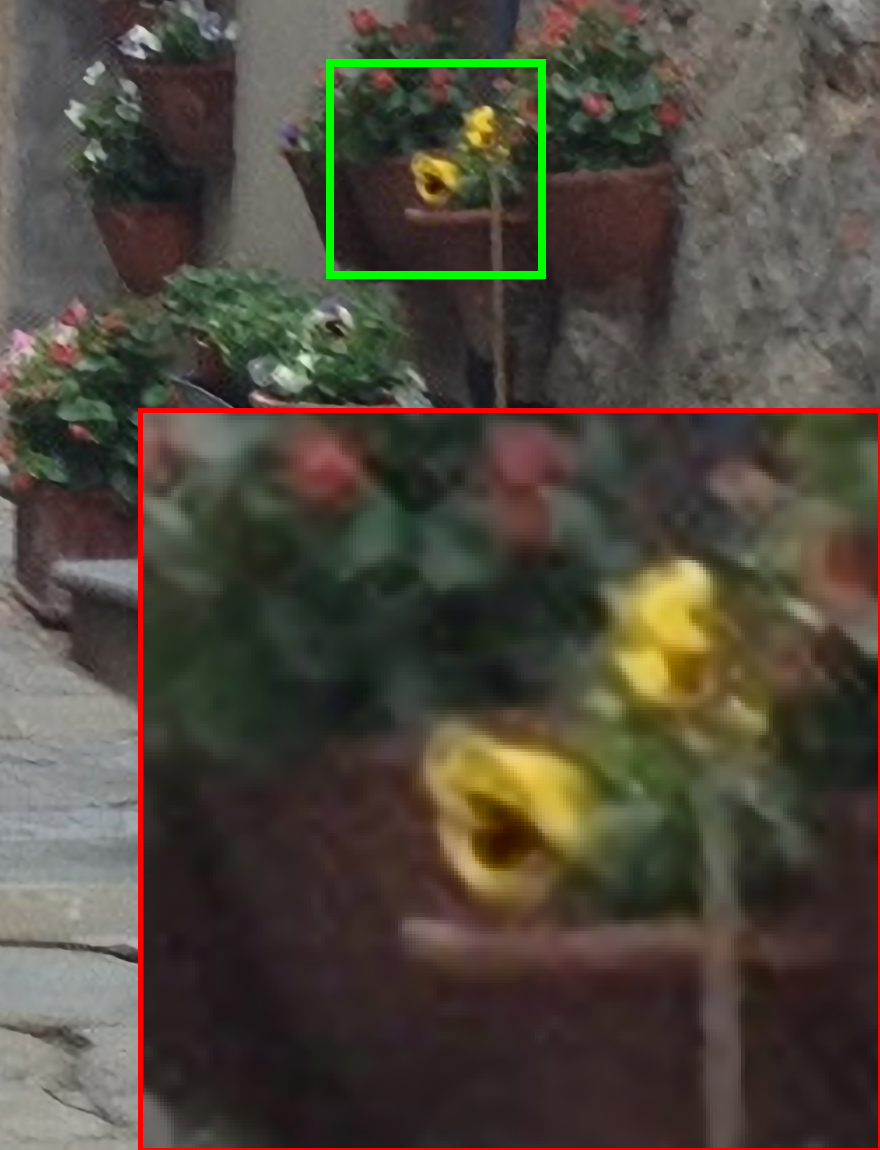}
			\caption*{MZSR (Ours) \\ \emph{10 updates}}
		\end{subfigure}
		
	\end{center}
	\caption{Visualized comparisons of super-resolution results ($\times 4$) on real-world image.}
	\label{fig:supp005}
\end{figure*}

To show the effectiveness of our MZSR, we visualize some results including scenarios with synthetic blur kernels and real-world images.
\figurename{~\ref{fig:supp001}} and \ref{fig:supp003} are the results on synthetic blur kernels. \ref{fig:supp005} is the result on a real-world image.

\end{document}